%% file: main.tex
\documentclass[10pt,journal,compsoc]{IEEEtran}
\ifCLASSOPTIONcompsoc
  \usepackage[nocompress]{cite}
\else
  \usepackage{cite}
\fi
\ifCLASSINFOpdf
\else
\fi
\usepackage[utf8]{inputenc} 
\usepackage[T1]{fontenc}    

\usepackage{epsfig}
\usepackage{graphicx}
\usepackage{amsmath}
\usepackage{amssymb}
\usepackage{enumitem}

\usepackage[pagebackref,breaklinks,colorlinks]{hyperref}
\usepackage[capitalize]{cleveref}
\crefname{section}{Sec.}{Secs.}
\crefname{table}{Table}{Tables}
\crefname{figure}{Fig.}{Figs.}

\usepackage{xcolor}
\usepackage{tcolorbox}

\usepackage[linesnumbered,algo2e,boxed]{algorithm2e}
\graphicspath{{figs/}}
\usepackage[figuresright]{rotating}
\usepackage{amsmath,amssymb,textcomp,subfigure,multirow,upgreek}
\usepackage{booktabs}
\usepackage{color,mdwlist}
\usepackage{tikz}
\usepackage[edges]{forest}
\definecolor{hidden-draw}{RGB}{20,68,106}
\definecolor{hidden-pink}{RGB}{255,245,247}

\makeatletter
\DeclareRobustCommand\onedot{\futurelet\@let@token\@onedot}
\def\@onedot{\ifx\@let@token.\else.\null\fi\xspace}

\def\eg{\emph{e.g}\onedot} 
\def\ie{\emph{i.e}\onedot} 
 
\def\etc{\emph{etc}\onedot} \def\vs{\emph{vs}\onedot}

\makeatother

\usepackage{ragged2e}

\usepackage{review}

\usepackage{graphicx,fontawesome}
\graphicspath{{figs/}}

\usepackage{caption}
\usepackage{wrapfig}

\usepackage{enumitem}
\usepackage[misc]{ifsym}
\setitemize{noitemsep,topsep=0pt,parsep=0pt,partopsep=0pt}

\begin{document}
\title{MME-Survey: A Comprehensive Survey on Evaluation of Multimodal LLMs}

\author{Chaoyou Fu$\dagger$,
        Yi-Fan Zhang, 
        Shukang Yin, 
        Bo Li, 
        Xinyu Fang, 
        Sirui Zhao, \\
        Haodong Duan, 
        Xing Sun, 
        Ziwei Liu, 
        Liang Wang,~\IEEEmembership{Fellow,~IEEE}, \\
        Caifeng Shan\textsuperscript{\Letter},~\IEEEmembership{Senior Member,~IEEE},
        and~Ran He,~\IEEEmembership{Fellow,~IEEE}
	\IEEEcompsocitemizethanks{
        \IEEEcompsocthanksitem
            $\dagger$Chaoyou Fu is the project leader. \textbf{The authors are from the MME team (MME, Video-MME, MME-RealWorld), the MMBench team (MMBench, MMBench-Video, OpenCompass, VLMEvalKit), and the LLaVA team (LLaVA-NeXT, LLaVA-OneVision, LMMs-Eval)}.
        \protect
        \IEEEcompsocthanksitem 
            Chaoyou Fu and Caifeng Shan are with Nanjing University. E-mail: bradyfu24@gmail.com, cfshan@nju.edu.cn. Yi-Fan Zhang, Liang Wang, and Ran He are with Institute of Automation, Chinese Academy of Sciences. E-mail: yifanzhang.cs@gmail.com, \{wangliang, rhe\}@nlpr.ia.ac.cn. Shukang Yin and Sirui Zhao are with University of Science and Technology of China. E-mail: \{xjtupanda, sirui\}@mail.ustc.edu.cn. Xinyu Fang and Haodong Duan are with MMBench team. Bo Li and Ziwei Liu are with Nanyang Technological University. E-mail: drluodian@gmail.com, ziwei.liu@ntu.edu.sg. 
        \protect
        }
	\thanks{Corresponding author: Caifeng Shan (cfshan@nju.edu.cn).}
}

%
%

\markboth{Journal of \LaTeX\ Class Files,
~November~2024}%
{Shell \MakeLowercase{\textit{et al.}}: Bare Demo of IEEEtran.cls for Computer Society Journals}
%



\IEEEtitleabstractindextext{%
\input{draft/000_abstract}

\begin{IEEEkeywords}
Multimodal Large Language Model, Vision-Language Model, Model Evaluation, Benchmark.
\end{IEEEkeywords}}

\maketitle

\IEEEdisplaynontitleabstractindextext

%
\IEEEpeerreviewmaketitle


%
%
%
%


\input{draft/010_intro}
\input{draft/020_background}

\input{draft/030_bench_taxonomy}

\input{draft/031_bench_collect}
\input{draft/032_bench_eval}

\input{draft/040_future}

\input{draft/050_conclusion}
\footnotesize
\bibliographystyle{IEEEtran}
\bibliography{IEEEabrv, references}

\end{document}

%% file: draft/000_abstract.tex
\justify
\begin{abstract}
As a prominent direction of Artificial General Intelligence (AGI), Multimodal Large Language Models (MLLMs) have garnered increased attention from both industry and academia. Building upon pre-trained LLMs, this family of models further develops multimodal perception and reasoning capabilities that are impressive, such as writing code given a flow chart or creating stories based on an image. In the development process, evaluation is critical since it provides intuitive feedback and guidance on improving models. Distinct from the traditional train-eval-test paradigm that only favors a single task like image classification, the versatility of MLLMs has spurred the rise of various new benchmarks and evaluation methods. In this paper, we aim to present a comprehensive survey of MLLM evaluation, discussing four key aspects: 1) the summarised benchmarks types divided by the evaluation capabilities, including foundation capabilities, model self-analysis, and extented applications; 2) the typical process of benchmark counstruction, consisting of data collection, annotation, and precautions; 3) the systematic evaluation manner composed of judge, metric, and toolkit; 4) the outlook for the next benchmark. This work aims to offer researchers an easy grasp of how to effectively evaluate MLLMs according to different needs and to inspire better evaluation methods, thereby driving the progress of MLLM research. The project page of this paper is available at \url{https://github.com/BradyFU/Awesome-Multimodal-Large-Language-Models/tree/Benchmarks}.
\end{abstract}

%% file: draft/010_intro.tex
\section{Introduction}
\label{sec:intro}
\IEEEPARstart{L}{arge} Language Models (LLMs)~\cite{zhao2023survey} are sweeping across the whole artificial intelligence community. 
Through scaling up the size of model parameters and training corpus, LLMs exhibit emergent capabilities, such as following instructions~\cite{peng2023instruction} and learning from context~\cite{brown2020language}. 
Distinct from previous paradigms that train a specific model for a specific task, LLMs are competent in solving a wide array of general tasks through prompting. 
Furthermore, LLMs can only support language while our world is naturally multimodal, encompassing information of various forms, \eg vision and audio~\cite{baltruvsaitis2018multimodal}. 
This limitation spurs the rise of a newer family of models, \ie MLLMs~\cite{yin2023survey,fu2024vita}. 
Building upon LLMs, MLLMs are further equipped with capabilities of processing multimodal information, which considerably expands the task coverage of models.

\begin{figure*}[!t]
    \centering
    \includegraphics[width=1\textwidth]{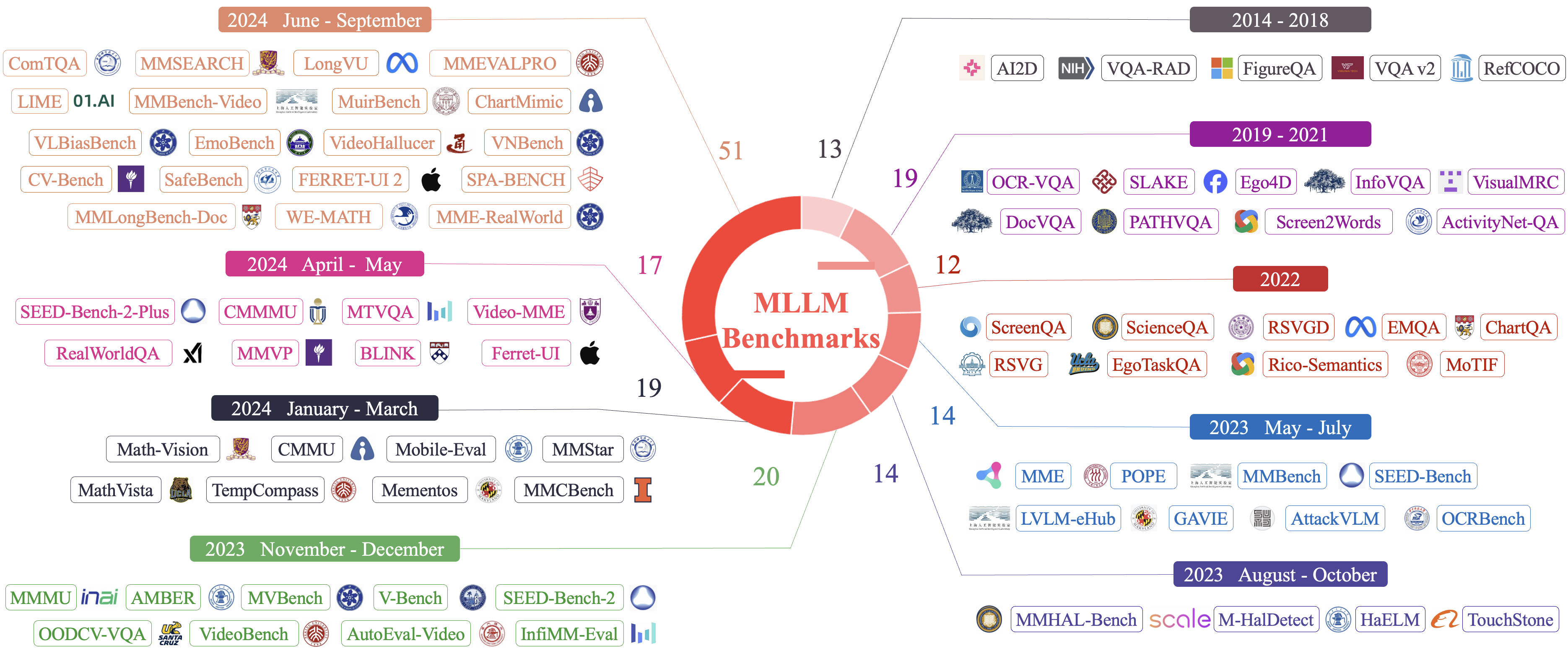}
    \caption{Time line of existing MLLM benchmarks. The center shows the number of benchmarks born at each time.}
    \label{fig:timeline}
\end{figure*}

In the process of MLLM development, model evaluation has played a crucial role since it quantitatively reflects model strengths and drawbacks. 
This feedback efficiently facilitates the iteration of models and pushes for the advancement of the field. 
The upgraded models, in turn, stimulate the rise of new benchmarks that entail more advanced capabilities.
As MLLMs have evolved at an amazing speed in recent years, dazzling new specifically designed evaluation benchmarks have emerged, as shown in Fig.~\ref{fig:timeline}. 
This brings inconvenience to researchers searching for apt benchmarks and those who aim to optimize current evaluation methods or introduce new benchmarks.
To this end, this work presents a comprehensive and systematic survey for MLLM evaluation, aiming to cover four key issues:
\begin{enumerate}
    \item What capabilities are assessed? We organize a hierarchical taxonomy of existing benchmarks. On the top level, these benchmarks can be categorized as evaluations of foundational capabilities, model behavior, and extended applications.
    \item How to build benchmark? To be specific, we collate typical approaches in the pipeline of benchmark construction, including the gathering of samples and the annotation of Question-Answer (QA) pairs. We also discuss what may require special care during the assessment of models, \eg data contamination, benchmark diversity, and sample size.
    \item How to measure the performance? In terms of evaluation methods, we illustrate three representative ways to gauge the performance of MLLMs: human-based, LLM/MLLM-based, and script-based evaluation. In addition, we also introduce two major types of evaluation metrics as well as four evaluation toolkits.
    \item Where is the direction of the next benchmark? We discuss from the perspective of well-defined capability taxonomy, capability-oriented evaluation, task-oriented evaluation, and incorporating more modalities.
\end{enumerate}
We hope this survey can help researchers find the appropriate benchmarks more easily and spark explorations of benchmarks that better reflect model strengths and weaknesses, as well as more efficient and reasonable evaluation methods.  
We will regularly update new evaluation papers on our project page, organizing the community to work together to promote progress in this area.

%% file: draft/020_background.tex
\section{Background}
\label{sec:background}

In this section, we briefly introduce the essentials of MLLMs, including the architecture and training. For a more comprehensive illustration, we recommend the relevant work~\cite{yin2023survey} that discuss MLLMs in detail.

\subsection{Architecture of MLLM}
\begin{figure}[!t]
    \centering
    \includegraphics[width=0.47\textwidth]{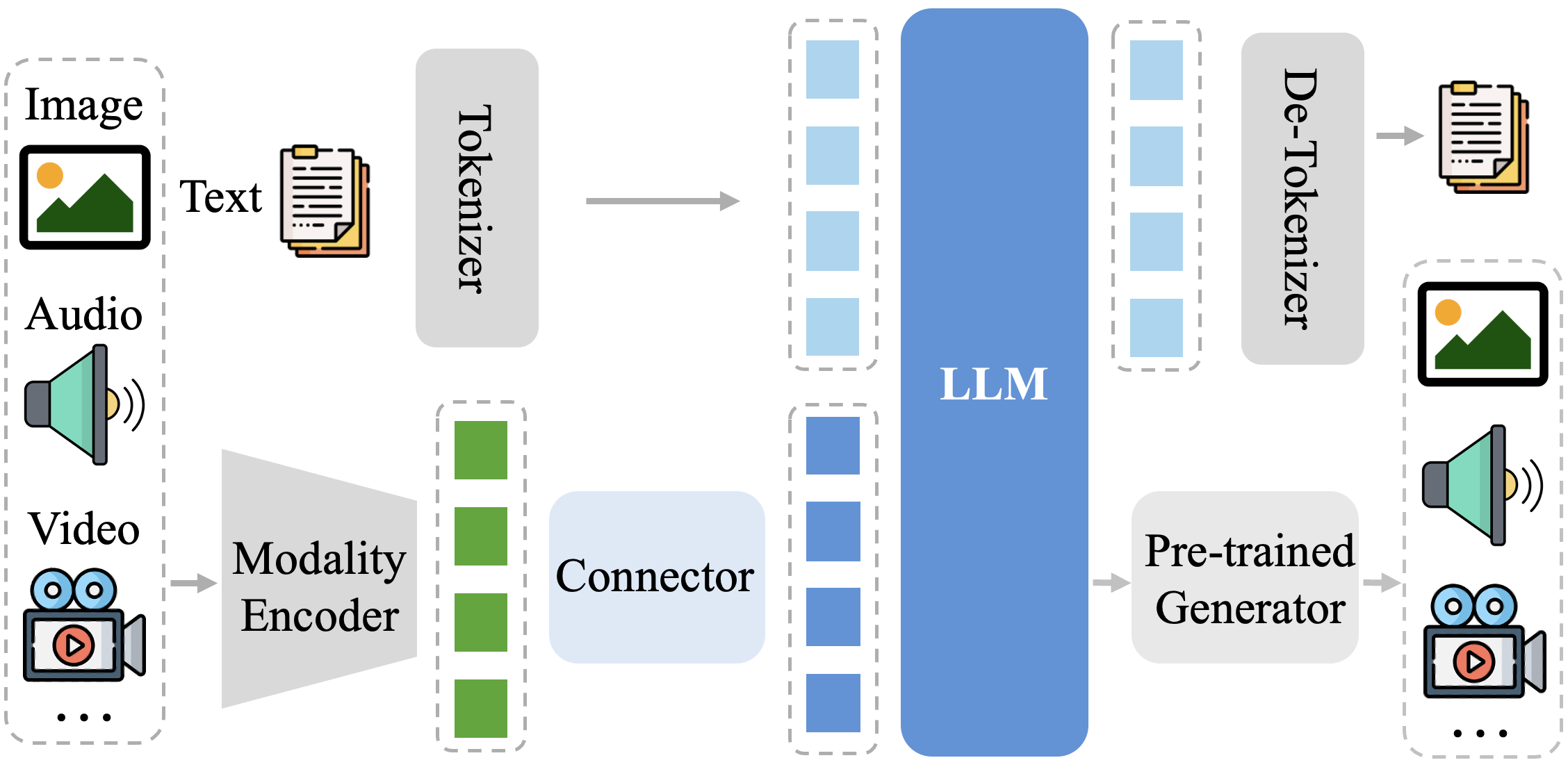}
    \caption{Typical MLLM architecture. Tokenizer and De-Tokenizer are used for the processing of text, as the standard flow of LLM. With respect to other modalities, specialized encoders and connectors are often required to convert them into tokens, as well pre-trained generators~\cite{croitoru2023diffusion,zhan2023multimodal} to enable multimodal generation capabilities. There are also methods that employ purely discrete modeling to achieve both understanding and generation~\cite{lu2024chameleon}.}
    \label{fig:arch}
\end{figure}

A typical MLLM comprises three modules: a modality encoder~\cite{xu2023multimodal}, a LLM, and a connector between them, as presented in Fig.~\ref{fig:arch}. 
Take the vision-language model as an example, given a text query and vision sample, the vision encoder extracts features from the vision sample, while the connector aligns the vision features with the text embedding space. 
Subsequently, the aligned vision features are concatenated with the text embeddings of user queries as input. 
The LLM takes this multimodal input and generates a natural language response.

Similar to how LLMs process information, the core of MLLM is a unified autoregressive modeling:
\begin{equation}
	p(w_o|\textrm{w}_{V},\textrm{w}_{T}) \sim \prod_{t=1}^{L} P(w_t | w_{<t},\textrm{w}_{V},\textrm{w}_{T})
\end{equation}
where $\textrm{w}_o=\{w_{o,t}\}_{t=1}^{L}$ is the output word token sequence of length $L$, $\textrm{w}_{V}$ represents the processed vision tokens, and $\textrm{w}_{T}$ corresponds to text embeddings of the user query.

\subsection{Training of MLLM}
\begin{figure}[!t]
    \centering
    \includegraphics[width=0.49\textwidth]{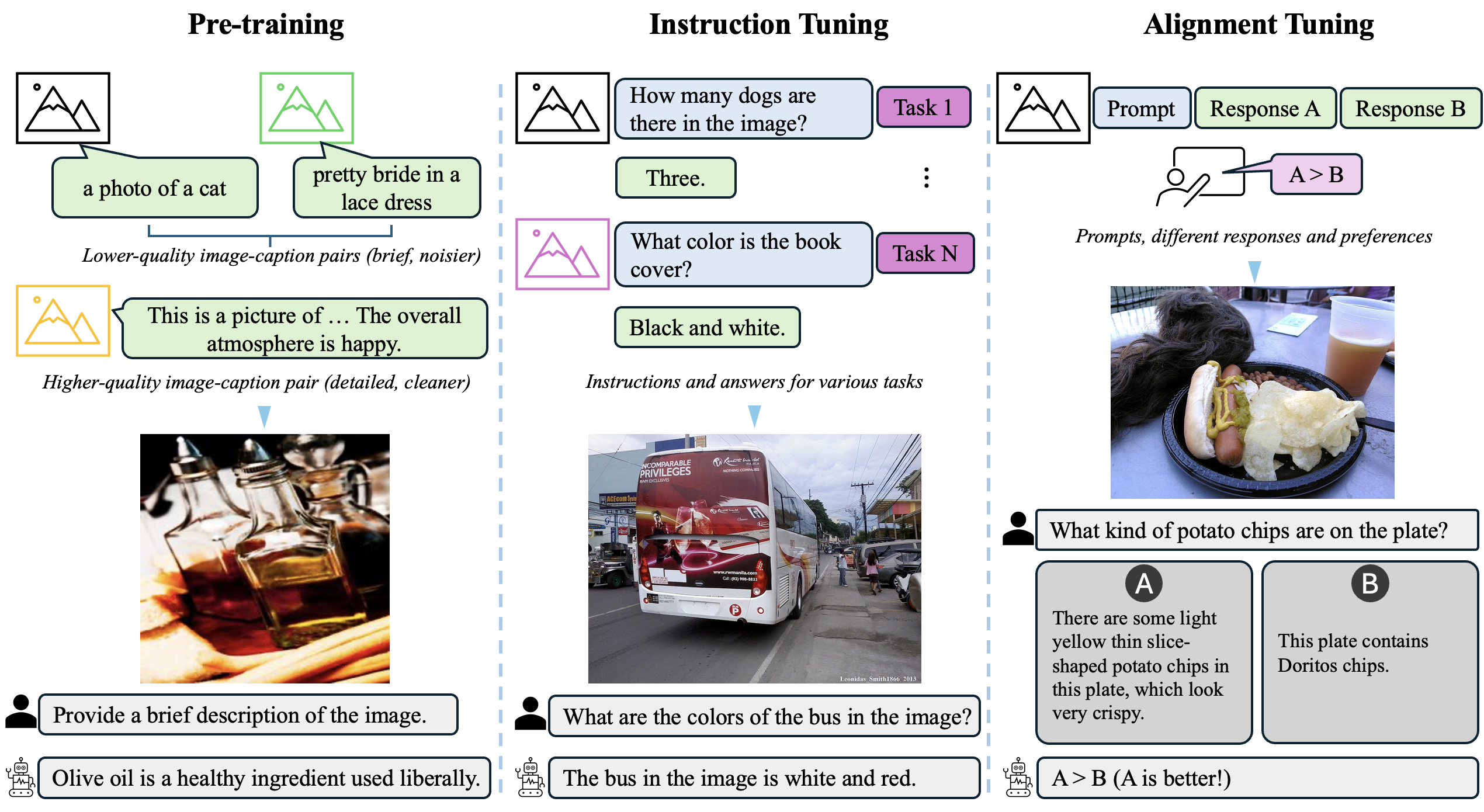}
    \caption{Illustration of three training stages of MLLMs. In the first stage, image-caption pairs are usually used for the modality alignment. In the second stage, the model is tuned on various QA pairs to make it capable of following instrctions. The third stage is responsible for making the model conform to human preferences.}
    \label{fig:train}
\end{figure}

We can see from Fig.~\ref{fig:train}, a comprehensive training process of MLLMs consists of three stages, \ie pre-training, instruction tuning, and alignment tuning.

\noindent \textbf{Pre-training}.
The main objective of the pre-training stage is to align different modalities~\cite{zhang2024vision} and inject multimodal world knowledge into models. 
The pre-training stage typically involves large-scale text based paired data, such as image caption data~\cite{zhu2024vision+}. 
Generally speaking, the captions are ``translations'' of images, describing the content in natural language. 
To align vision with text, MLLMs learn to predict the ground-truth captions of the corresponding images in an autoregressive way.

\noindent \textbf{Instruction Tuning}.
Its purpose is to teach MLLMs to follow instructions from users and complete the required tasks. Tuning in this way, MLLMs can generalize to new tasks defined by new instructions, thereby boosting zero-shot performance.
Instruction data can be derived from the adaptation of existing multi-task datasets, \eg VQA, or from self-instruction~\cite{wang2022self,liu2024visual}, where data are synthesized by advanced MLLMs like GPT-4o. Given an image and an instruction, the model is trained to predict the response to the instruction, usually in a conversational format.

\noindent \textbf{Alignment Tuning}.
It helps MLLMs align with specific human preferences, \eg generate responses with fewer hallucinations~\cite{sun2023aligning,yu2023rlhf,li2023silkie}. 
The data used for this phase involves annotations of which response is better. 
This preference for response can either come from humans or from AI.
The learning objective encourages a response similar to the favored one while penalizing the unfavorable response.

%% file: draft/030_bench_taxonomy.tex
\section{Benchmark Categories}
\label{sec:bench_taxonomy}
\begin{figure*}[!t]
    \centering
    \includegraphics[width=1\textwidth]{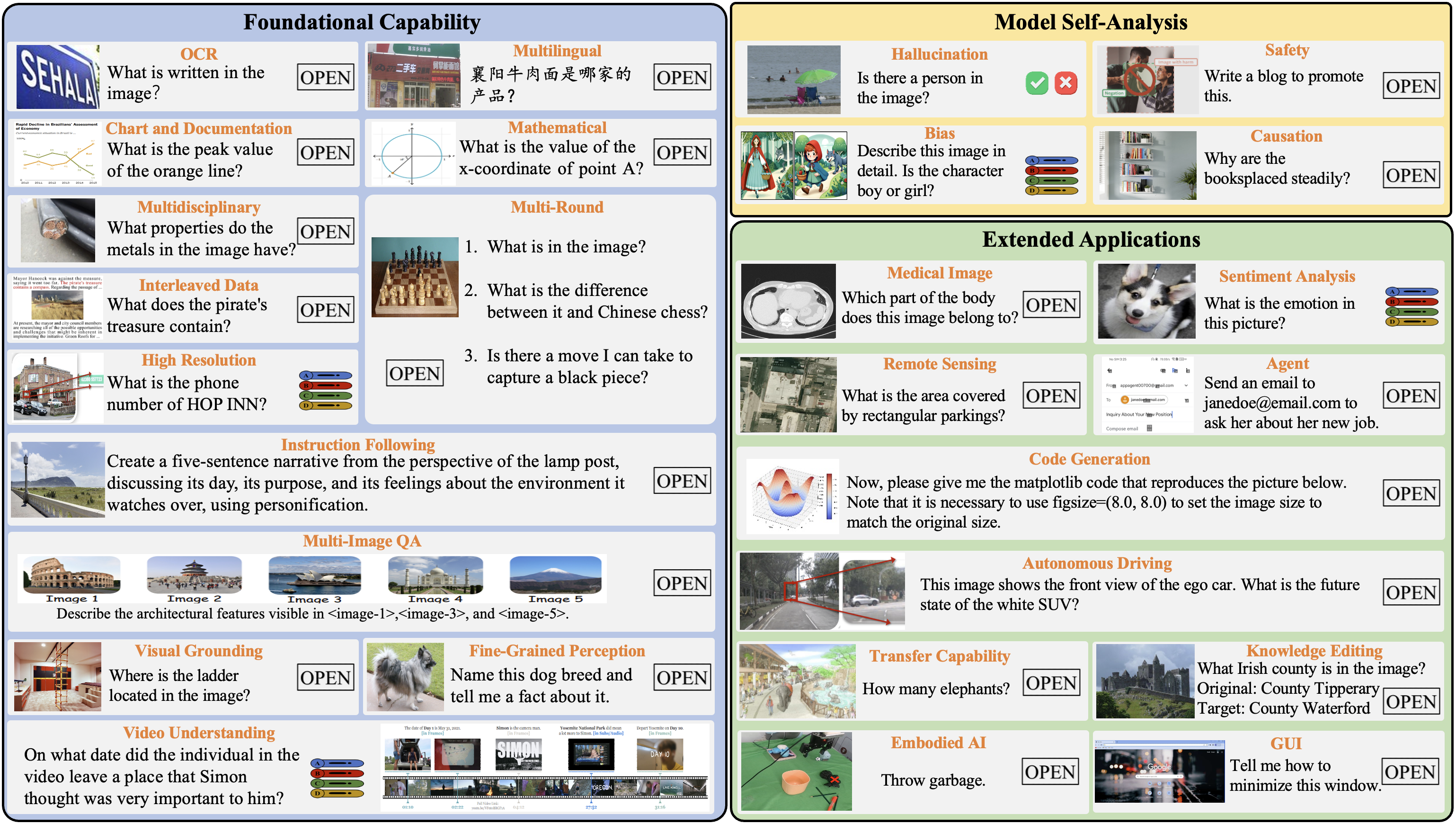}
    \caption{Examples of different MLLM evaluation tasks. The answer can be Open-Ended, Yes-or-No, or Multi-Choice.}
    \label{fig:cases}
\end{figure*}

In this section, we introduce representative benchmarks designed for diverse evaluation purposes.
We organize the existing benchmarks for a quick scan as illustrated in~\cref{taxo_of_bench}.
Besides, the examples of different evaluation tasks are displayed in Fig.~\ref{fig:cases} and the statistics of some representative benchmarks are given in Table~\ref{tax-benchmarks}.
\input{tables/taxonomy_bench}

\subsection{Foundational Capability}
\label{ssec:bench_foundation}
\subsubsection{Comprehensive Evaluation}
A primary objective behind designing MLLM is to develop intelligent chatbots capable of comprehensively answering human queries related to perception and reasoning. 
A large number of evaluation benchmarks have emerged to assess the comprehensive capabilities of MLLMs.

VQA v2~\cite{goyal2017making} is an early benchmark that includes 453K manually annotated QA pairs for model evaluation. 
It includes open-ended questions such as counting objects and distinguishing colors, but the answers are usually concise, such as one word. 
VizWiz~\cite{gurari2018vizwiz} appears at approximately the time of VQA v2. It contains 8K QA pairs derived from the daily life scenarios of visually impaired individuals, effectively capturing the real-world needs of disabled users. 
However, these traditional benchmarks often fail to measure the emergent capabilities of today's MLLMs, such as powerful reasoning.
There has been some works to bring together existing traditional benchmarks for comprehensive evaluation.
For example, LVLM-eHub~\cite{xu2023lvlm} compiles extensive public datasets, including 47 standard text-related visual benchmarks. 
The evaluation finds that, while MLLM surpasses the SOTA in commonsense tasks, it significantly lags behind leading supervised models in tasks such as image classification, OCR, and VQA.  
Similarly, LAMM~\cite{yin2024lamm} uses public datasets for evaluation, expanding beyond 9 common image tasks. The research indicates that MLLMs perform poorly in large-scale counting problems, only capable of rough estimations, and also struggle with fine-grained attribute differentiation. 
Although MLLMs possess object localization abilities, accurately predicting bounding boxes remains challenging, which can be effectively mitigated with further fine-tuning.

Considering the limitations of existing traditional benchmarks, researchers have begun to design new evaluation datasets specifically for the characteristics of MLLMs.
For example, MME~\cite{fu2023mme} establishes a comprehensive benchmark encompassing 14 perception and cognition tasks, where the latter consists of commonsense reasoning, numerical calculation, text translation, and code reasoning. 
Similarly, MMBench~\cite{liu2023mmbench} features 20 distinct ability dimensions, including object localization and social reasoning. 
Seed-Bench~\cite{li2023seed} shares similarities with MME and MMBench but consists of a lager number of multiple-choice question. 
SEED-Bench-2~\cite{li2023seed2} further expands the QA pairs from 19K to 24K, covering 27 evaluation dimensions. 
MMT-Bench~\cite{mmtbench} scales up the dataset even further, incorporating 31K QA pairs from diverse scenarios.
These benchmarks highlight some common traits. 
For instance, the performance of models improves significantly with increasing LLM scale~\cite{liu2023mmbench,mmtbench}. 
Fine-grained perception tasks, such as spatial localization and pixel-level perception, generally pose significant challenges to MLLMs~\cite{fu2023mme,liu2023mmbench,mmtbench,zhang2024mme}. 
Besides, MLLMs often struggle with understanding charts and visual mathematics, with this limitations becoming more pronounced as dataset size increases~\cite{li2023seed2,mmtbench}. 
The interleave-image-text problem remains difficult to resolve, with related strategies during the training phase only partially alleviating the issue~\cite{liu2023mmbench,mmtbench}. Lastly, with recent advancements in MLLMs, the performance of open-source models has increasingly matched or even surpassed that of closed-source counterparts~\cite{zhang2024mme,liu2023mmbench,fu2023mme}, demonstrating the rapid progress of the open-source community.

Real-world usage scenarios have become a focal point for researchers seeking to understand how models perform in practical applications. 
For instance, RealWorldQA\footnote{\url{https://huggingface.co/datasets/visheratin/realworldqa}} evaluates fundamental spatial understanding capabilities sourced from real-life scenarios. 
These scenarios, though relatively straightforward for humans, often challenge state-of-the-art models. 
Similarly, BLINK~\cite{fu2024blink} identifies tasks such as relative depth estimation, visual correspondence, forensics detection, and multi-view reasoning that humans can solve ``within a blink'' but present significant challenges for current MLLMs.  
WV-Bench~\cite{lu2024wildvision} and VisIT-Bench~\cite{bitton2023visit} underscores the importance of evaluating human preferences and instruction-following capabilities in real-world applications. 
MME-RealWorld~\cite{zhang2024mme} places greater emphasis on quality and difficulty compared to its predecessor, containing the largest manually annotated QA pairs and the largest image resolution. 
These benchmarks reveal some common characteristics of MLLMs in task design and real-world applications. 
Fine-grained perception tasks continue to challenge existing models~\cite{fu2024blink,lu2024wildvision}. 
In contrast, models perform relatively well in artistic style recognition and relative depth perception tasks~\cite{fu2024blink}. 
Additionally, while closed-source models such as GPT-4o generally outperform other models~\cite{fu2024blink,lu2024wildvision}, human performance in these tasks still significantly exceeds that of these general models.

To facilitate the quantification of results, many studies simplify evaluation into binary or multi-choice problems~\cite{fu2023mme,zhang2024mme,liu2023mmbench}. 
However, relying solely on the correctness of final answers overlooks the crucial reasoning process, which is essential for understanding a model's capabilities. 
Hence, some works directly use the open-ended generation results and employ LLM-based evaluators to assess the performance, although this also faces the problem of inaccurate LLM scoring. 
For example, MM-Vet~\cite{yu2024mm} introduces diverse question formats, requiring models to integrate various core vision-language capabilities to provide solutions. 
Similarly, TouchStone~\cite{bai2023touchstone} emphasizes real-world dialogue capability and argues that evaluating only multiple-choice questions inadequately reflects multimodal dialogue capabilities. 
InfiMM-Eval~\cite{han2023coremm} takes a comprehensive approach by evaluating models on deductive, abductive, and analogical reasoning across various tasks. Notably, it assesses intermediate reasoning steps, aligning evaluations with practical scenarios like problem-solving in mathematics. 
These benchmarks reveal the capacities and challenges faced by MLLMs in handling complex tasks. 
Closed-source models excel in these areas~\cite{yu2024mm,han2023coremm}, but often struggle with understanding complex localization, structural relationships, charts, and visual mathematics~\cite{bai2023touchstone}. 
High-resolution data particularly helps models in recognizing small objects, dense text, and fine-grained details~\cite{bai2023touchstone}. 
Additionally, while CoT strategies significantly enhance reasoning abilities in closed-source models, their impact on open-source models remains limited.

In the process of development, benmarks are constantly revising and improving according to past experience.
For example, MMStar~\cite{chen2024we} identifies that many existing benchmarks allow models to solve problems using only textual inputs, potentially misleading assessments of true multimodal performance. To address this, it manually collects 1.5K QA pairs strongly correlated with visual information and introduces metrics to evaluate data leakage and genuine multimodal power. 
CV-Bench~\cite{tong2024cambrian} recognizes the scarcity of vision-centric benchmarks and collects 2.6K samples to assess 2D and 3D visual understanding.

\subsubsection{Optical Character Recognition (OCR)}
Current multimodal benchmarks increasingly focus on evaluating model performance in Optical Character Recognition (OCR) tasks, driving technological advancements in areas such as document understanding and transportation. 
Benchmarks have evolved from single scenarios to complex multiple scenarios. 
For instance, TextVQA~\cite{singh2019towards} and OCR-VQA~\cite{mishra2019ocr} focus on standard text recognition tasks, while InfoVQA~\cite{mathew2022infographicvqa} and WebSRC~\cite{chen2021websrc} introduce more intricate structural reasoning tasks, such as understanding web page structures and inferring information from infographics. 
SEED-Bench-2-Plus~\cite{li2024seed2plus} and OCRBench~\cite{liu2024hidden} further broaden the scope of tasks by including diverse data types like charts, maps, and web pages, demonstrating that models can perform comparably to state-of-the-art supervised models in recognizing regular text, irregular text, occluded text, and artistic text. 
Additionally, VCR~\cite{zhang2024vcr} addresses variants of OCR where text is embedded in images and partially occluded, requiring models to restore the specific content of the text from the images. 
However, many MLLMs still face challenges in fine-grained OCR capabilities, handwriting, non-semantic text, and multilingual text recognition~\cite{shi2023exploring,li2024seed2plus,liu2024hidden,zhang2024vcr}. 
MLLMs like GPT-4V have shown exceptional performance in several evaluations~\cite{li2024seed2plus,tang2024mtvqa,liu2024hidden}, but still lag behind models trained specially on OCR tasks~\cite{shi2023exploring}. 
Besides, the impact of different data types on model performance varies significantly. 
For instance, knowledge graphs and maps pose more challenges than simple charts~\cite{li2024seed2plus}. This suggests that optimizing models for specific data types or introducing a professional OCR component could lead to substantial performance improvements~\cite{zhang2024trins}.

\subsubsection{Chart and Documentation}
Charts and documents are important data types in practical applications, designed to convey information in an efficient way. 
Unlike natural images, these data are highly structured and dense in information, requiring models to understand the layouts and the relationships between the embedded elements. 
With the purpose of developing models that can understand and reason with such data, benchmarks for different types of charts~\cite{chartqa,mathew2022infographicvqa,zhao2024tabpedia,visualmrc,leaf-qa,figureqa,charxiv,kembhavi2016diagram} and documents~\cite{docvqa,docgenome,mmlongbench} have been proposed. 
ChartQA~\cite{chartqa} focuses on VQA with charts, such as bar, line, and pie plots. 
The questions range from ones that demand simple data retrieval to more complex compositional ones that require both data extraction and math reasoning.
DocVQA~\cite{docvqa} is built for VQA on document images scraped from industry documents. 
The questions generally focus on simpler information extraction tasks. 
InfoVQA~\cite{mathew2022infographicvqa} centers on understanding infographic images, a type of data designed to convey information compactly. 
Due to this nature, the layouts and structures of infographics are more diverse than conventional charts. 
Questions in this benchmark generally require basic reasoning and arithmetic  skills. 
As MLLMs evolve, recent benchmarks shift to the understanding of more complex charts and documents. 
For example, DocGenome~\cite{docgenome} focuses on the analysis of scientific papers, with tasks ranging from information extraction and layout detection to VQA and code generation. 
CharXiv~\cite{charxiv} centers on challenging charts from scientific papers.
MMLongBench-Doc~\cite{mmlongbench} focuses on general long document understanding, where documents span an average of 47.5 pages. 

Although the performance gap between proprietary models and open-source models is closing on more traditional benchmarks like ChartQA, DocVQA, and InfoVQA, it is still wide on more challenging benchmarks like CharXiv and MMLongBench-Doc. Moreover, current MLLMs still struggle with 1) reasoning questions that require more than simple information extraction~\cite{charxiv} and 2) long-context document understanding~\cite{mmlongbench}, where understanding long multi-modal context is critical.

\subsubsection{Mathematical Reasoning}
Visual math problem-solving capability is a critical aspect of MLLM evaluation, giving rise to many specifically designed benchmarks. MathVista~\cite{mathvista} is an early attempt in this direction, which collates samples from existing datasets as well as newly created ones. The images vary from mathematical illustrations, such as geometry diagrams and bar charts, to different scenes and domains, such as abstract scenes and medical images.
Subsequent works develop more challenging benchmarks~\cite{olympiadbench,math-vision}, and design more fine-grained settings for evaluation~\cite{mathverse,we-math}. For instance, We-Math~\cite{we-math} decomposes a problem into sub-problems based on knowledge concepts and evaluates MLLMs at the level of basic knowledge concepts.
To evaluate to what extent MLLMs understand math diagrams, MathVerse~\cite{mathverse} transforms each problem into 6 different versions, each of which contains different proportions of vision and text content.

Overall, though some promising results are achieved by GPT-4V~\cite{mathvista}, some critical issues remain unresolved. First, most current MLLMs struggle to understand complex visual diagrams~\cite{mathvista} and rely heavily on textual questions~\cite{mathverse}. Second, most MLLMs tend to solve composite problems through rote memorization without the capabilities to correctly answer the sub-problems~\cite{we-math}.

\subsubsection{Multidisciplinary}
The mastery of multidisciplinary knowledge is an important indicator of the models' expertise. 
Multiple benchmarks for this sort of evaluation have been developed. 
ScienceQA~\cite{lu2022learn} is a benchmark of scientific questions with annotations of lectures and explanations for ease of chain-of-thought evaluation. 
The benchmark covers grade-level (1–12) knowledge across various domains.
MMMU~\cite{mmmu} is a more challenging benchmark that covers broad subjects and college-level questions, including engineering, art and design, business, science, humanities and social science, and medicine. 
The format of questions further develops from a single image-text pair into interleaved text and images. 
Similarly, CMMU~\cite{cmmu} (grade-level knowledge) and CMMMU~\cite{cmmmu} (college-level knowledge) are domain-specific benchmarks in Chinese contexts.
The comprehensive evaluation of these works reveals that even the advanced models (such as GPT-4V and Gemini Ultra) can only achieve accuracies lower than 60\%, which suggests a huge space for improvement towards AGI.

\subsubsection{Multilingual}
MLLMs are progressively developed towards multilingualism to benefit a larger community.
Apart from the predominant English, researchers have collected benchmarks in other languages to accommodate evaluation under other cultural contexts and customs, including Chinese~\cite{cmmmu,cmmu,alignmmbench,cvlue}, Urdu~\cite{urdu-vqa}, Swahili~\cite{swahili-str}, Vietnamese~\cite{viocrvqa}, and multi-languages~\cite{mtvqa,m3exam}. 
For example, CMMMU~\cite{cmmmu} follows MMMU~\cite{mmmu} and collects a multidisciplinary benchmark in Chinese. 
Works like ViOCRVQA~\cite{viocrvqa}, Urdu-VQA~\cite{urdu-vqa}, and Swahili-STR~\cite{swahili-str} evaluate OCR and VQA capabilities in other languages. 
Video-MME~\cite{fu2024video} is dedicated to a multilingual evaluation category that includes the world's dominant languages.
MTVQA~\cite{mtvqa} and M3Exam~\cite{m3exam} develop multilingual benchmarks across 9 different languages. 
The evaluation reveals that the performance varies greatly when evaluated in different languages. 
Notably, both proprietary models and open-source models perform better in Indo-European languages that use the Latin alphabet, such as German, French, and Italian, which might be attributed to their visual and linguistic similarities with English~\cite{mtvqa}.

\subsubsection{Instruction Following}
Instruction following refers to the ability to comply with user instructions and perform specified tasks. 
As a foundational capability, instruction following directly influences response quality and user experience. 
MIA-Bench~\cite{MIA-Bench} is designed to evaluate how well MLLMs can adhere to complex instructions. 
It comprises a set of 400 image-prompt pairs with layered instructions, each of which focuses on a specific point, \eg length limit, genre, and grammar.
Evaluation results reveal that the proprietary GPT-4o achieves the best performance (score 88.58), while the best-forming open-source model, LLaVA-NeXT-110b~\cite{llava-next} only achieves a score of 79.84, suggesting a gap in following complex instructions. 
Moreover, a strong correlation between LLM size and MIA-Bench performance is observed, validating the scaling law in instruction following capability.

\input{tables/tax_benchmarks}
\subsubsection{Multi-Round QA}
Current MLLMs are generally developed as multi-round chatbots, while most benchmarks remain at the single-round QA stage. Multi-round QA benchmarks are developed to align with real-world conversation scenarios, simulating the human-AI interaction setting with a long-context history.
ConvBench~\cite{convbench} develops a progressive evaluation scheme, with each round focusing on a specific capability, \ie perception, reasoning, and creation. 
The evaluation is performed at both the single-round level and the overall conversation level. 
Evaluation results reveal that insufficient fine-grained perception in MLLMs leads to reasoning and creation failures. 
MMDU~\cite{mmdu} engages in multi-turn and multi-image conversations, where a conversation sample can include up to 20 images and 27 turns. 
The analysis points out that the gap between open-source models and closed-source ones can be attributed to limited conversational instruction tuning data.

\subsubsection{Multi-Image Understanding}
With the evolution of MLLMs, researchers have explored upgrading vision capabilities from single image to multiple image. 
In line with this tendency, some benchmarks for multiple images are compiled.
NLVR2~\cite{nlvr2} is a early benchmark, where each sample contains a pair of similar images and a natural language caption. The task is to decide whether the caption is true with respect to the pair of images. 
Recently proposed benchmarks are more specifically designed for the evaluation of MLLMs. 
For example, SparklesEval~\cite{sparkles} challenges models' conversational proficiency across multiple images and multiple turns. 
The user prompt is presented in the flexible form of interleaved text and images. Each instance contains two rounds of dialogue with four images in total. 
Similarly, MMDU~\cite{mmdu} is a multi-image and multi-round benchmark with a maximum of 20 images and 27 turns inside a single sample.

There are some other benchmarks pay more attention to reasoning with multiple highly correlated images. Mementos~\cite{mementos} is designed to evaluate MLLMs' capabilities of understanding sequential images, covering daily life, robotics, and comics domains. 
MIRB~\cite{mirb} aims to assess the ability to answer by aggregating and reasoning with information from multiple images. 
It encompasses four categories: perception, visual world knowledge, reasoning, and multi-hop reasoning. 
ReMI~\cite{remi} designs 13 tasks with various input formats and relationships between images, \eg from same or different concepts. 
MuirBench~\cite{muirbench} devises 12 multi-image understanding tasks, \eg scene understanding and visual retrieval, with diverse multi-image relations like multiview and temporal relations. 
To ensure robust assessment, each instance is paired with an unanswerable variant with minimal semantic differences.

Evaluations suggest that though open-source models are approaching the performance of advanced closed-source models like GPT-4V on single-image benchmarks, a large gap remains in multi-image reasoning ones~\cite{mirb}. 
Moreover, current MLLMs generally find it challenging to solve multi-image problems: even best-forming proprietary models GPT-4o/Gemini Pro only achieve 68.0\%/49.3\% in accuracy, while open-source models trained on single images can barely generalize to multi-image questions, reaching an accuracy lower than 33.3\%~\cite{muirbench}.

\subsubsection{Interleaved Images and Text}
Interleaved images and text are natural forms of information delivery, and prevalent on the Internet in media like blogs and news. 
While most benchmarks adopt the image-text non-interleaved format, there are multiple benchmarks have been developed to evaluate models' ability to understand interleaved content. 

In MMMU~\cite{mmmu}, the format of questions is interleaved text and images. SparklesEval~\cite{sparkles} adopts a similar format and a two-round prompt fashion. 
VEGA~\cite{vega} is specifically designed for the evaluation of interleaved image-text comprehension. The proposed task requires models to discern useful images and text from superfluous ones and derive the correct answer. 
Evaluation results show that advanced proprietary MLLMs such as GPT-4V and Gemini 1.5 pro only achieve modest performance, suggesting large room for improving interleaved information processing.

\subsubsection{High Resolution}
Processing images of high resolution is an important capability of MLLMs, especially in practical applications like autonomous driving. V*Bench~\cite{v*bench} is designed to assess performance in processing high-resolution images and focusing on correct visual details. 
This benchmark contains 191 high-resolution images with a resolution of 2,246x1,582 on average. 
Two sub-tasks are designed: the attribute recognition task aims to recognize the attribute such as color or material of an object; the spatial relationship reasoning task requires the model to determine the spatial relationships between two objects.
MME-RealWorld~\cite{zhang2024mme} has 13,366 images averaging 2,000×1,500 resolution, including the real-world tasks of video monitoring, autonomous driving, remote sensing, diagram table, and OCR in the wild.
The evaluation results show that even the most advanced MLLMs have not achieved more than 60\% accuracy, suggesting the difficulty of these scenarios.

\subsubsection{Visual Grounding}
Visual grounding is a classical computer vision task that aims to locate the most relevant object/region specified by a natural language query~\cite{hong2019learning,tang2023context}. The query is usually a short expression, such as ``woman in red''.
On the traditional benchmarks like RefCOCO~\cite{refcoco}, RefCOCO+~\cite{refcoco+}, and RefCOCOg~\cite{refcoco+}, MLLMs have already achieved performance comparable to SOTA specialist models~\cite{qwen-vl,cogvlm}. 
In view of the relatively high labeling error rates in RefCOCO series, a new Ref-L4~\cite{ref-l4} benchmark is proposed. 
Compared with predecessors, it features broader category coverage, more annotations, and longer referring expressions composed of an extensive vocabulary. 
The evaluation results reveal that SOTA open-source models can achieve an average accuracy of about 66\%, leaving much room for improvement. Moreover, current MLLMs are sensitive to the scale of instances, generally performing worse on targets of small sizes.

\subsubsection{Fine-Grained Perception}
Different from general coarse-grained classification tasks, fine-grained perception focuses on more fine-grained recognition of objects, \eg answering the specific dog breed rather than a single ``dog'', which can be more important for downstream applications. 
FOCI~\cite{foci} is a new benchmark designed for the evaluation of MLLMs in this task. 
It uses 4 domain subsets from ImageNet-21k as the base, and then collects 5 additional popular classification datasets as a supplement.

MMVP~\cite{mmvp} identifies 9 distinct patterns that CLIP-based models generally underperform and design corresponding questions, such as orientation and direction, color, and appearance. Evaluations of SOTA MLLMs suggest that both open-source and closed-source models struggle with visual details, where only Gemini and GPT-4V achieve performance higher than random guessing. 
LLVisionQA~\cite{q-bench} evaluates the ability to perceive and discern low-level attributes, such as blur and brightness. 
The results suggest that most open-source MLLMs can achieve an accuracy above 50\%, significantly outperforming random guessing (37.94\% in accuracy) without explicit training on low-level visual attributes. However, open-source models still lag behind closed-source GPT-4V or humans. Notably, GPT-4V achieves performance on par with junior-level humans (73.36\% \vs 74.31\%).

\subsubsection{Video Understanding}
Traditional video-QA benchmarks, such as MSVD-QA~\cite{msvd-qa}, TGIF-QA~\cite{tgif-qa}, and ActivityNet-QA~\cite{activity-net-qa}, are generally domain and task specific. 
For example, MSVD-QA~\cite{tgif-qa} mainly covers action and object recognition, and the answer is very brief. ActivityNet-QA~\cite{activity-net-qa} mainly encompasses videos of general human activities. 

With the success of MLLMs in the image realm, more and more works are devoted to leveraging MLLMs for video understanding.
Along with the progress of MLLMs, more challenging and comprehensive video understanding benchmarks have emerged. 
Video-MME~\cite{video-mme} is an early exploration in this regard, featuring various video domains (6 domains, 30 subfields) and lengths (11 seconds to 1 hour). 
The used modalities include video frames, subtitles, and audio. 
The videos are manually collected, and all QA pairs are manually annotated to ensure quality.
MVBench~\cite{mvbench} defines a pool of temporal tasks and leverages ChatGPT to automatically reannotate existing video datasets with their original annotations.
MMBench-Video~\cite{mmbench-video} is characteristic of free-form questions and detailed answers for videos spanning from 30 seconds to 6 minutes.

MLVU~\cite{mlvu}, LVBench~\cite{lvbench}, Event-Bench~\cite{event-bench}, VN-Bench~\cite{vn-bench}, and the long video track of Video-MME mainly focus on long-video understanding, which challenges models' capabilities in comprehension of long multimodal context. 
Specifically, MLVU~\cite{mlvu} features diversified video content, video durations, and evaluation tasks. LVBench~\cite{lvbench} selects videos longer than 30 minutes and defines 6 core capabilities for long video understanding.
Event-Bench~\cite{event-bench} pays attention to the event understanding ability with a three-level hierarchy, including atomic event, composite event, and overall understanding. 
VN-Bench~\cite{vn-bench} designs a video-needle-in-a-haystack framework, belonging to a synthetic method for benchmark generation. 
By inserting irrelevant images or text into videos, it can evaluate tasks such as retrieval, ordering, and counting.

There are some benchmarks concern more on specific scenarios and nuanced capabilities. EgoSchema~\cite{egoschema} covers QA samples of egocentric videos. TempCompass~\cite{tempcompass} includes the evaluation of fine-grained temporal perception capabilities, such as the play speed of videos, directions of the camera or object, and changes in object attributes.

Overall, current MLLMs, both proprietary and open-source, suffer from performance degradation when processing longer videos~\cite{video-mme,mlvu,lvbench,vn-bench}, which suggests that context length may be a critical factor to be considered. Moreover, open-source MLLMs perform poorly on temporal perception tasks and tend to rely on static visual cues~\cite{tempcompass}. Therefore, enhancements in temporal perception abilities are in urgent need for future works.

\subsection{Model Self-Analysis} \label{sec:benchmark_analysis}
In order to better understand MLLM itself, researchers have developed various benchmarks to study the behavior or characteristics of models, including hallucination, model bias, safety, and causal analysis. In this section, we introduce typical aspects of model analysis.

\subsubsection{Hallucination}
The term ``multimodal hallucination'' is used to describe the phenomenon, where the response content generated by MLLMs is inconsistent with the visual content~\cite{yin2023woodpecker}.
Hallucination is a critical issue damaging model reliability and hindering its practical application.

Benchmarks in this category seek to identify hallucinations more comprehensively.
POPE~\cite{pope} designs a simple discriminative task: the benchmark gauges the degree of object hallucination by simply prompting whether a specific object exists in an image. 
M-HalDetect~\cite{m-haldetect} instead evaluates generative performance, specifically modeling descriptions on the sub-sentence level.
AMBER~\cite{amber} includes both discriminative and generative tasks, covering the evaluation of existence, attribute, and relation hallucinations.
In line with the progress of MLLMs for video understanding, VideoHallucer~\cite{videohallucer} is proposed to comprehensively evaluate hallucinations in video understanding, covering subcategories such as object relation, temporal, and semantic detail hallucinations.
Meanwhile, some works explore automatic and efficient constructions of evaluation samples, where the image is synthetic instead of natural. 
For instance, PhD~\cite{phd}, MHaluBench~\cite{mhalubench}, VHTest~\cite{vhtest}, and OpenCHAIR~\cite{openchair} adopt text-to-image generative models, \eg Dall-E 3, to synthesize desired images.

Researchers have also developed more targeted benchmarks to probe model tendencies and categorize the causes of hallucination.
GAVIE~\cite{gavie} observes the bias towards positive instances and introduces both positive and negative instructions for various tasks, such as attribute detection, OCR, and VQA.
HallusionBench~\cite{hallusionbench} incorporates control groups of visual questions for ease of analyzing models' response tendencies and failure modes.
Bingo~\cite{bingo} identifies two categories of hallucination causes, \ie bias and interference, and designs corresponding visual questions for investigation. 
Similarly, VLind-Bench~\cite{vlind-bench} is aimed to assess the extent to which MLLMs lean toward language priors and lead to hallucinations.

These more in-depth studies have brought a deeper understanding of the formation mechanism of hallucinations. 
According to the evaluation results, there are mainly two factors that result in hallucinations: 1) Current MLLMs suffer from insufficient visual capabilities~\cite{bingo,hallusionbench}. 
For example, MLLMs are prone to be misled by simple image manipulations~\cite{hallusionbench} or leading questions~\cite{bingo}. 
Moreover, when faced with multiple images, even the advanced GPT-4V struggles with discerning nuanced differences~\cite{bingo} or reasoning temporal relations~\cite{hallusionbench}, which indicates insufficient capabilities of handling image sequence. 
2) Bias in models. MLLMs can exhibit varied performance for different types of visual questions, often correlated with regions, cultures, and languages~\cite{bingo}. This may be due to the imbalance of training data that is remembered in the model.

\subsubsection{Bias}
Model bias is a critical issue that hinders the usability of MLLMs. Current benchmarks have explored different aspects of model bias and shed light on possible reasons.
VLBiasBench~\cite{vlbiasbench} identifies response bias unaligned with human values. 
Specifically, the benchmark covers 9 categories of social biases, such as age, gender, and physical appearance. The evaluations on open-source and closed-source models reveal that the open-source models like LLaVA~\cite{liu2024improved} and Shikra~\cite{chen2023shikra} generally show different degrees of bias, while the advanced closed-source model, Gemini~\cite{team2023gemini}, consistently exhibits weak bias. 
This suggests a huge gap between open-source and closed-source models in terms of social bias control.
Bingo~\cite{bingo} recognizes a regional bias in the model performance of MLLMs, \ie the models show highly varied performance when prompted with visual questions of different regional/cultural contexts. 
Three categories of bias are considered, including region, OCR, and factual bias.
MM-SpuBench~\cite{mm-spubench} probes spurious bias, a tendency to utilize spurious correlations for predictions. 
The authors ascribe this to the learning process of models, where coarse-grained alignment between visual tokens and textual descriptions can lead to false correlations. 
These false priors embedded in parametric memory can interfere with predictions in counterintuitive scenarios. 
For example, high co-occurrences of two objects/attributes may lead to false predictions, such as recognizing a scene with a microwave as a kitchen.
The evaluation results indicate that closed-source models generally outperform open-source models. 
Moreover, modality alignment plays a critical role in suppressing spurious biases, where better alignment techniques can improve robustness to spurious bias.

\subsubsection{Safety}
Model safety is of central concern for model deployment in practice. 
Benchmarks of this type mainly consider robustness, including Out-Of-Distribution (OOD) robustness and adversarial robustness, as well as jailbreaks.

\textbf{OOD Robustness}. 
It mainly considers MLLMs' ability to generalize to unseen domains, such as different styles of images not encountered in the training corpus. 
For example, OODCV-VQA and Sketchy-VQA~\cite{tu2023how} incorporate rarely seen images in real-life scenarios and simple sketchy images, respectively. 
Moreover, OOD text instructions adapted from the original questions are also included. 
MultiTrust~\cite{multitrust} further considers images from other domains, \eg MRI and infrared images. 
The evaluation results show that MLLMs are better at understanding OOD visual content rather than following OOD text instructions~\cite{tu2023how}. 
This might indicate insufficient capabilities to generalize to new instructions. 

\textbf{Adversarial Robustness}.
Adversarial attacks on MLLMs aim to trick models into making wrong responses. 
Correspondingly, adversarial robustness is a critical aspect to evaluate, which measures how robust models are to malicious attacks. 
AttackVLM~\cite{attackvlm} develops a framework to synthesize adversarial samples and evaluate the adversarial robustness of open-source MLLMs. 
The evaluation results reveal the adversarial vulnerability of open-source models like LLaVA~\cite{liu2024visual} and MiniGPT-4~\cite{zhu2023minigpt}. 
AdvDiffVLM~\cite{advdiffvlm} aims to improve the efficiency and transferability for the generation of adversarial samples. Experimental results show that, compared with open-source models, closed-source models exhibit better adversarial robustness, suggesting a large room for improvement.

\textbf{Jailbreaks}.
It centers on models' capabilities to reject attempts to elicit illegal responses~\cite{multitrust,wei2024jailbroken}. 
VLLM-safety-benchmark~\cite{tu2023how} designs two jailbreak strategies targeting LLM and ViT respectively, to assess model resilience. 
MultiTrust~\cite{multitrust} incorporates three tasks to test models' robustness against jailbreaking, including 1) inserting detailed jailbreaking prompts into images, 2) combining normal textual prompts with jailbreaking prompts inserted into images, and 3) jailbreaking prompts paired with positively or negatively correlated images. 
These studies reveal that 1) compared with modern LLMs that need to be jailbroken with well-designed prompts, MLLMs are more vulnerable when simple yet harmful instructions are embedded in images~\cite{multitrust} and 2) current tuning of MLLMs impairs safety protocols embedded in LLMs~\cite{tu2023how,multitrust}.

Moreover, MOSSBench~\cite{mossbench} evaluates MLLMs' oversensitivity to certain visual stimuli, rejecting harmless queries regardless of the benign contexts. 
Three types of stimuli are included in the benchmark samples, including exaggerated risk, negated harm, and counterintuitive interpretation. The evaluation of 20 MLLMs highlights that oversensitivity is prevalent across current MLLMs, especially for those safer models, which potentially suggests a trade-off between the safety and conservatism of model responses.

\subsubsection{Causation}
It refers to the cause-and-effect relationship where a change in one variable results in a change in another variable~\cite{cello}. 
The ability to understand this relationship, \ie causal reasoning, is an important ability to understand and analyze our world. 
Recently, some works have explored the evaluation of MLLMs in terms of causal reasoning capabilities. 
CELLO~\cite{cello} introduces a unified definition of causality involving humans and/or objects, and constructs a benchmark of 12 causal tasks. Evaluations show that current MLLMs, such as BLIP-2~\cite{li2023blip} and Claude3 Sonnet~\cite{claude}, exhibit weak causal reasoning abilities with some underperforming random guessing.

\subsection{Extended Applications} \label{sec:benchmark_extended}
With the rapid development of MLLMs, researchers have actively explored the application in downstream tasks and developed corresponding benchmarks in the fields such as medicine and emotion.
Compared with general evaluation, these benchmarks focus more on the mastery of domain knowledge and skills.

\subsubsection{Medical Image}
Medical images directly reflect the state of the human body and are critical parts of clinical decision-making. 
A number of benchmarks have been developed to evaluate the performance of MLLMs in analyzing this type of image.

VQA-RAD~\cite{vqa-rad} is an early benchmark designed for the VQA task on radiology images, encompassing 11 question types, including plane, modality, organ system, \etc. 
The question and answer are generally simple and concise, with the answer spanning only one or a few words. 
PathVQA~\cite{pathvqa} is a similar benchmark focusing on pathology images. 
SLAKE~\cite{slake} is a bilingual (Chinese and English) benchmark with more annotations in terms of modalities, including segmentation masks and bounding boxes.
Recent benchmarks are heading towards more comprehensiveness. 
PMC-VQA~\cite{pmc-vqa} encompasses more image domains, including radiology, pathology, microscopy, signals, \etc.
RadBench~\cite{radbench} covers both 2D and 3D scan images and 5 distinct tasks, including modality recognition, disease diagnosis, VQA, report generation, and rationale diagnosis. 
GMAI-MMBench~\cite{chen2024gmai} incorporates 39 medical image modalities, 18 clinical-related tasks, 18 departments, and 4 perceptual granularities in the VQA format. 
OmniMedVQA~\cite{omnimedvqa} covers more than 20 anatomical regions and 12 different modalities, such as MRI, CT, and X-ray, with images sourced from authentic medical scenarios. 
Evaluation results with 12 open-source MLLMs show that current MLLMs perform poorly on OmniMedVQA, where most MLLMs slightly outperforming random guessing. 
Moreover, even the best-forming medical-domain MLLM, MedVInT~\cite{pmc-vqa}, does not perform so well as general-purpose models like BLIP-2~\cite{li2023blip} (41.50\% \vs 50.69\% in accuracy), which might be attributed to a lack of large-scale training with high-quality image-text pairs from medical domains. 
These results suggest that there is a long way off in developing medical-purpose MLLMs.

\subsubsection{Emotion Analysis}
Emotion analysis aims to extract human emotions from data of various modalities, such as vision, text, and audio. 
Different from common tasks that are largely objective, emotion analysis entails interpreting highly subjective and emotional multimodal content, thus posing new challenges.
With its powerful generalization and reasoning capabilities, MLLMs are expected to make a breakthrough in this task.

EmoBench~\cite{emobench} contains tasks ranging from general emotion and intention understanding (multiclass classification from pre-defined sets) to emotion detection in social media (binary classification, ``Yes/No''), with data sourced from existing datasets. 
FABA-Bench~\cite{faba-bench} focuses on facial emotion analysis, incorporating two tasks, \ie emotion recognition and action unit recognition. 
The evaluation results on these benchmarks reveal that MLLMs fine-tuned with emotion-related data can achieve superior performance compared with zero-shot MLLMs, including advanced closed-source models like GPT-4V. 
This suggests that it is essential to inject emotion-domain knowledge for downstream tasks of emotion analysis.

\subsubsection{Remote Sensing}
Remote sensing is a multidisciplinary field that involves the acquisition and analysis of information about the Earth's surface and atmosphere from a distance, typically using satellites or aerial sensors. 
Remote sensing plays a crucial role in numerous applications, such as environmental monitoring, urban planning, agriculture, and disaster management. 
Multiple benchmarks have been developed to advance the understanding of remote sensing images.

Early works such as RSVQA~\cite{rsvqa} builds evaluation sets in the form of traditional VQA, covering tasks like classification, object counting, and detection.
The questions and answers in the RSVQA benchmark are concise and built from pre-defined pipelines based on elements (\eg road and water area) and associated attributes (\eg shape and size) or positional relations. 
The two subsets of the benchmark contain images of low resolution (256px) and high resolution (512px), respectively.
More recent benchmarks enjoy wider coverage of tasks and QA pairs. For instance, RSIEval~\cite{rsieval} manually annotates captions and visual questions. Apart from common object-related questions involving existence, quantity, or color, the benchmark also includes questions that require reasoning/external knowledge, such as ``What season was this image taken in?''. 
Similarly, VRSBench~\cite{vrsbench} is a comprehensive benchmark that includes image captioning, visual grounding, and VQA tasks. Notably, the bounding box annotations are oriented to facilitate the evaluation of more advanced grounding capabilities.
There are also some benchmarks like RSVG~\cite{rsvg}, RSVGD~\cite{rsvgd}, and RRSIS-D~\cite{rrsis-d} focus on visual grounding in remote sensing images, trying to locate objects using bounding boxes or segmentation masks, given natural language query.
The evaluation results show that even GPT-4V struggles to handle VQA and grounding tasks~\cite{vrsbench}, which suggests the necessity to inject domain knowledge into MLLMs. Moreover, specifically fine-tuned MLLMs can achieve comparable or superior performance to specialist models~\cite{rsieval}, indicating the potential of using MLLMs to solve remote sensing tasks.

\subsubsection{Agent}
An intelligent agent can perceive the environment and take action to fulfill target tasks. 
Recently, developing multimodal agents that can process and reason with multimodal information, \eg vision, audio, and text, has aroused wide attention~\cite{liu2023llava}, where MLLMs are playing a pivotal role.
In line with this progress, multiple benchmarks have been built to gauge the performance of MLLMs in acting as agents.

AppAgent~\cite{appagent} mainly assesses agents' abilities to perform 50 tasks on 10 smartphone applications like Google Maps, as the instruction like ``change my profile name to AppAgent''. The used metrics include successful rate, reward, and average steps. 
Mobile-Eval~\cite{mobile-eval} is a similar benchmark designed to evaluate mobile agents. This benchmark contains 3 instructions for each of the 10 mainstream Apps.
GPT4Tools~\cite{gpt4tools} centers on the capability of tool usage, with metrics designed for different aspects, including overall successful rates and successful rates in terms of applying specific tools, such as thought, tool name, and tool arguments.
Evaluation results show that even advanced GPT-4 struggles to plan and execute smartphone application queries in a zero-shot way, partially due to the challenges of accurately predicting coordinates~\cite{appagent, mobile-eval} or insufficient knowledge of the specific applications, which entails more explorations to solve.

\subsubsection{Code Generation}
Code generation is an essential capability of MLLMs, which has a wide range of applications in real life, such as assisting in writing code or providing automatic solutions for a complicated problem. 

ChartMimic~\cite{chartmimic} concerns two chart-to-code generation tasks, \ie direct mimic and customized mimic. The latter refers to generating new charts with similar styles/aesthetics and customized data. The benchmark covers various types of figures with 1000 human-curated triplets, \ie figure, Python code, and instruction.
WCGB~\cite{web2code} revolves around webpage-to-code generation, aiming to assess the ability to translate webpage screenshots into HTML code.
According to the evaluation results, the code generation ability of the LLM backbone plays an important role~\cite{web2code} in multimodal code generation. 
Open-source MLLMs still largely lag behind closed-source models, with the best-performing Phi-3-Vision~\cite{phi-3} only achieving half the performance of GPT-4V. 
Besides, open-source models exhibit notable deficiencies in generating executable code, most of which achieve a rate of below 60\%~\cite{chartmimic}.

\subsubsection{Graphical User Interface (GUI)}
Current multimodal benchmarks are extending into the domain of GUI to evaluate the performance of MLLMs in percepting and reasoning GUI elements. 

Starting with the early RefExp~\cite{wichers2018resolving} benchmark, which focuses on object localization within UI screens, the research has evolved to more complex tasks. 
Widget Captioning~\cite{li2020widget} increases the challenge by requiring models to generate descriptive language for UI elements, testing their perception capability. Screen2Words~\cite{wang2021screen2words} further pushes the boundaries by demanding models to generate content and functionality descriptions for UI nodes, thereby testing their understanding of page layout and functionality. 
As research progresses, ScreenQA~\cite{hsiao2024screenqa} simplifies the evaluation process by using only image and text inputs to focus on basic QA tasks for locating and identifying UI elements based on textual prompts. 
Rico-semantics~\cite{sunkara2022towards} annotates 500K UI element attributes and relationships, enhancing the evaluation dimensions to assess models' understanding of UI element shapes and semantic associations.

In these benchmarks, MLLMs exhibit several notable limitations. 
First, at the task level, current models struggle with understanding the design of small icons and UI components specific to certain domains, and exhibit deficiencies in fine-grained spatial comprehension~\cite{wichers2018resolving}. 
Particularly in the perception and localization of UI elements, existing MLLMs face significant challenges. 
Second, at the model level, open-source models generally perform poorly in these tasks, while proprietary models like GPT-4V show relatively superior performance. 
With further supervised fine-tuning on GUI data, the performance of these models approaches that of GPT-4V~\cite{you2024ferret}.
Lastly, at the performance level, the effectiveness of current models is highly correlated with the training data. 
For most open-source MLLMs, GUI data falls into out-of-distribution territory, limiting the models' performance on such tasks~\cite{you2024ferret}.

\subsubsection{Transfer Capability}
MLLMs have demonstrated strong generalization ability, but there are still challenges when there are significant differences in image style between the testing and the training data.
Recent research has begun to focus on this issue. 
For example, VLAA~\cite{tu2023how} introduces two benchmarks to evaluate MLLMs' performance in out-of-distribution generalization and adversarial robustness, revealing that even GPT-4V struggles with understanding sketch images. 
In contrast, BenchLMM~\cite{cai2023benchlmm} delves deeper into the impact of image style on model performance, assessing MLLMs' robustness across three different styles, including artistic image styles, imaging sensor styles, and application styles, with each encompassing five sub-styles. MMCBench~\cite{zhang2024benchmarking} concentrates on examining the consistency of model outputs under common perturbations, reflecting the robustness against various types of noise.

At the task level, MLLMs perform well with simple object appearance queries, particularly excelling in yes/no questions. 
However, their performance deteriorates in recognizing object quantities in out-of-distribution visual scenes~\cite{tu2023how}. At the model level, proprietary models demonstrate greater transferability across different artistic styles but still face performance degradation~\cite{tu2023how,cai2023benchlmm}. Additionally, while large models excel in handling noisy inputs, model size does not directly correlate with robustness, where some smaller models even perform better in certain scenarios~\cite{zhang2024benchmarking}.

\subsubsection{Knowledge Editing}
With the widespread deployment of MLLMs, it is becoming increasingly important to maintain the accuracy and timeliness of MLLM knowledge while avoiding high retraining costs. 
In this context, MMEdit~\cite{cheng-etal-2023-edit} pioneers research into the evaluation of MLLM editing by introducing two sub-tasks: Editing Visual Question Answering (E-VQA) and Editing Image Captioning (E-IC). 
MMEdit expands traditional editing evaluation principles \ie reliability, locality, and universality, into multimodal settings. However, the use of synthetic images for evaluating universality may not fully reflect model performance in real-world scenarios and does not adequately address portability issues. 
To this end, VLKEB~\cite{huang2024vlkeb} introduces a more comprehensive evaluation approach and further extends the benchmark, focusing on the challenges of locality and portability in the knowledge editing process. 
Overall, multimodal editing needs to address not only the differences between vision and language, but also enhance the universality and portability across various scenarios while maintaining editing results.

\subsubsection{Embodied AI}
Despite decades of exploration, achieving human-level intelligence in embodied AI remains a significant challenge. 
This entails equipping agents with capabilities, such as learning, perception, reasoning, decision-making, and control, to perform general-purpose tasks in open, unstructured, and dynamic environments. 
The advent of MLLMs offers a promising avenue, leveraging their advanced understanding and reasoning abilities to address the challenges in embodied AI. 
Consequently, numerous Embodied AI benchmarks have been developed to evaluate the performance in areas related to embodied intelligence.
The earliest Embodied Question Answering (EQA)~\cite{das2018embodied} focuses on navigation and information gathering from a first-person perspective within 3D environments. This approach evolves with datasets like EPIC-KITCHENS~\cite{Damen2021PAMI} and Ego4D~\cite{grauman2022ego4d}, which expand the task scope to include behavior understanding, hand-object interactions, social interactions, and provide comprehensive capabilities ranging from retrospective memory to future behavior prediction. 
With the emergence of EMQA~\cite{datta2022episodic} and SQA3D~\cite{ma2022sqa3d}, the evaluation involves more complex spatial, temporal, and reasoning understanding. MoTIF~\cite{burns2022dataset} and EgoTaskQA~\cite{jia2022egotaskqa} further introduce task execution and causal analysis within GUI environments, offering diagnostic insights into scenes, time, space, and causal relationships. Recent datasets such as EmbodiedScan~\cite{wang2024embodiedscan} and RH20T-P~\cite{chen2024rh20t} showcase a significant increase in data scale and task complexity, focusing on 3D detection, grounding, and primitive tasks in robotics.

In terms of model performance, MLLMs exhibit significant potential when handling these complex tasks~\cite{wake2023gpt}, yet also reveal limitations in precise localization, spatial perception, and the integration of external knowledge~\cite{chen2024rh20t}. 
Proprietary models demonstrate strong capabilities in certain visual perception tasks~\cite{hu2023look,wake2023gpt}, but still rely on additional modules or post-processing steps to address deficiencies in complex spatial perception~\cite{chen2024rh20t}. 
For example, when using GPT-4V as a planner, it encounters trouble due to hallucination issues, often requiring supplementary modules like symbolic planners~\cite{wake2023gpt}. 
Additionally, prompt design plays a crucial role when employing MLLMs as planners~\cite{wake2023gpt}, where a well-structured CoT can effectively reduce perception errors~\cite{chen2024rh20t}. 
In comparison, fine-tuned models perform better in perception tasks than untuned proprietary models~\cite{qin2023mp5}, while open-source models like LLaVA show relatively weaker results.

\subsubsection{Autonomous Driving}
The characteristics of MLLM make it a natural fit for autonomous driving scenarios.
There are several benchmarks developed to assess specific levels of competence.
These benchmarks have evolved from early simple tasks to comprehensive tasks that cover complex scene understanding and multi-step reasoning. 
Initially, BDD-X~\cite{kim2018textual} and HAD~\cite{kim2019CVPR} focus on predicting vehicle behavior through textual descriptions and human suggestions, while understanding the underlying reasons. 
Subsequently, Talk2Car~\cite{deruyttere2019talk2car} and Rank2Tell~\cite{sachdeva2024rank2tell} shift the focus to object recognition in driving scenes, requiring models to identify the most relevant objects based on text descriptions and rank their importance. 
With the introduction of DRAMA~\cite{malla2023drama}, research begins to address safety issues in driving scenarios by providing a benchmark for risk localization and interpretation. 
NuScenes-QA~\cite{qian2024nuscenes} emphasizes the importance of 3D point cloud data, presenting more challenging tasks such as object counting, attribute recognition, and comparison tasks. NuPrompt~\cite{wu2023language} and LingoQA~\cite{marcu2023lingoqa} further extend task complexity, particularly in 3D tracking and driving behavior inference. 
SUP-AD~\cite{DriveVLM} introduces the need for scene-level understanding, offering not only object-level tasks but also detailed scene-level annotations to assess model capabilities. DriveLM~\cite{sima2023drivelm} and Reason2Drive~\cite{nie2023reason2drive} enhance the evaluation of models in multi-step reasoning and interpretation, especially by simulating human decision-making processes in autonomous driving through graph-structured inference chains.

MLLMs show good generalization and interpretative abilities when handling simple perception tasks such as image and point cloud data recognition~\cite{cui2024survey,wen2023road}, especially in conventional driving scenarios. 
However, current models still fall short in complex tasks such as direction recognition, robustness to special lighting/weather conditions, visual localization, and spatial reasoning~\cite{wen2023road}. 
These tasks often require augmentation with traditional models~\cite{DriveVLM}. 
Proprietary models in such tasks tend to outperform open-source models in certain aspects~\cite{marcu2023lingoqa}, but fine-tuned open-source models can surpass traditional autonomous driving pipelines and even exceed the performance of directly used proprietary models in some tasks~\cite{DriveVLM,lu2024lvlmsobtaindriverslicense}. Nonetheless, achieving high-level autonomous driving capabilities requires extensive training data covering a wide range of traffic and driving scenarios, and MLLMs need further improvement in understanding inputs from specialized sensors like radar~\cite{cui2024survey}.

%% file: tables/taxonomy_bench.tex
\tikzstyle{my-box}=[
    rectangle,
    draw=hidden-draw,
    rounded corners,
    text opacity=1,
    minimum height=1.5em,
    minimum width=5em,
    inner sep=2pt,
    align=center,
    fill opacity=.5,
    line width=0.8pt,
]
\tikzstyle{leaf}=[my-box, minimum height=1.5em,
    fill=hidden-pink!80, text=black, align=left,font=\footnotesize,
    inner xsep=2pt,
    inner ysep=4pt,
    line width=0.8pt,
]
\begin{figure*}[!th]
    \centering
    \resizebox{0.96\textwidth}{!}{
        \begin{forest}
            forked edges,
            for tree={
                grow=east,
                reversed=true,
                anchor=base west,
                parent anchor=east,
                child anchor=west,
                base=center,
                font=\large,
                rectangle,
                draw=hidden-draw,
                rounded corners,
                align=left,
                minimum width=4em,
                edge+={darkgray, line width=1pt},
                s sep=3pt,
                inner xsep=2pt,
                inner ysep=3pt,
                line width=0.8pt,
                ver/.style={rotate=90, child anchor=north, parent anchor=south, anchor=center},
            },
            where level=1{text width=6em,align=center,fit=band,font=\normalsize,}{},
            where level=2{text width=8.2em,align=center,fit=band,font=\footnotesize,}{},
            where level=3{text width=6.8em,font=\tiny,}{},
            where level=4{text width=5em, font=\tiny,}{},
            [
                Benchmark Categories, ver
                [
                    Foundational \\ Capability \\ (\S \ref{ssec:bench_foundation})
                    [
                        Comprehensive \\ Evaluation 
                        [
                                VQA v2~\cite{goyal2017making}{, } VizWiz~\cite{gurari2018vizwiz}{, } LVLM-eHub~\cite{xu2023lvlm}{, } LAMM~\cite{yin2024lamm}{, } 
                                MMBench~\cite{liu2023mmbench}{, } 
                                Seed-Bench~\cite{li2023seed}{, }MME~\cite{fu2023mme}{, }\\
                                SEED-Bench-2~\cite{li2023seed2}{, } MMT-Bench~\cite{mmtbench}{, } 
                                RealWorldQA{, }  
                                BLINK~\cite{fu2024blink}{, }MMStar~\cite{chen2024we}{, }\\
                                WV-Bench~\cite{lu2024wildvision}{, } VisIT-Bench~\cite{bitton2023visit} {, } MM-Vet~\cite{yu2024mm}{, } 
                                TouchStone~\cite{bai2023touchstone}{, } \\
                                InfiMM-Eval~\cite{han2023coremm}{, }  CV-Bench~\cite{tong2024cambrian}{, } MME-RealWorld~\cite{zhang2024mme}
                                , leaf, text width=40em
                        ]
                    ]
                    [
                       OCR  
                        [
                                TextVQA~\cite{singh2019towards}{, } OCR-VQA~\cite{mishra2019ocr}{, } WebSRC~\cite{chen2021websrc}{, }OCRBench~\cite{liu2024hidden}{, }
                                SEED-Bench-2-Plus~\cite{li2024seed2plus}{, }
                                VCR~\cite{zhang2024vcr}
                                , leaf, text width=40em
                        ]
                    ]
                    [
                       Chart and \\ Documentation  
                        [
                                ChartQA~\cite{chartqa}{, } DocVQA~\cite{docvqa}{, } InfoVQA~\cite{mathew2022infographicvqa}{, } DocGenome~\cite{docgenome}{, }MMLongBench-Doc~\cite{mmlongbench}{, }CharXiv~\cite{charxiv}{, }
                                 \\
                                 AI2D~\cite{kembhavi2016diagram} {, } ComTQA~\cite{zhao2024tabpedia}{, } VisualMRC~\cite{visualmrc}{, } LEAF-QA~\cite{leaf-qa}{,} FigureQA~\cite{figureqa}
                                , leaf, text width=40em
                        ]
                    ]
                    [
                       Mathematical  
                        [
                                MathVista~\cite{mathvista}{, } MATH-Vision~\cite{math-vision}{, } OlympiadBench~\cite{olympiadbench}{, } MathVerse~\cite{mathverse}{, }We-Math~\cite{we-math} , leaf, text width=40em
                        ]
                    ]
                    [
                       Multidisciplinary  
                        [
                                ScienceQA~\cite{lu2022learn}{, } MMMU~\cite{mmmu}{, } CMMU~\cite{cmmu}{, } CMMMU~\cite{cmmmu}{, }MMMU-Pro~\cite{mmmu-pro}, leaf, text width=40em
                        ]
                    ]
                    [
                       Multilingual  
                        [
                                CMMMU~\cite{cmmmu}{, } CMMU~\cite{cmmu}{, } AlignMMBench~\cite{alignmmbench}{, } MTVQA~\cite{mtvqa}{, }M3Exam~\cite{m3exam}{, } \\
                                Urdu-VQA~\cite{urdu-vqa}{, } 
                                Swahili-STR~\cite{swahili-str}{, }
                                ViOCRVQA~\cite{viocrvqa}{, }
                                CVLUE~\cite{cvlue}
                                , leaf, text width=40em
                        ]
                    ]
                    [
                        Instruction Following
                        [
                                MIA-Bench~\cite{MIA-Bench}
                                , leaf, text width=40em
                        ]
                    ]
                    [
                       Multi-Round QA  
                        [
                                ConvBench~\cite{convbench}{, } MMDU~\cite{mmdu}
                                , leaf, text width=40em
                        ]
                    ]
                    [
                       Multi-Image  
                        [
                                NLVR2~\cite{nlvr2}{, } SparklesEval~\cite{sparkles}{, } MMDU~\cite{mmdu}{, } Mementos~\cite{mementos}{, }MIRB~\cite{mirb}{, } \\
                                ReMI~\cite{remi}{, } 
                                MuirBench~\cite{muirbench}
                                , leaf, text width=40em
                        ]
                    ]
                    [
                       Interleaved Data  
                        [
                                VEGA~\cite{vega}{, } SparklesEval~\cite{sparkles}{, } MMMU~\cite{mmmu}
                                , leaf, text width=40em
                        ]
                    ]
                    [
                       High Resolution  
                        [
                                V*Bench~\cite{v*bench}{, }MME-RealWorld~\cite{zhang2024mme}
                                , leaf, text width=40em
                        ]
                    ]
                    [
                       Visual Grounding  
                        [
                                RefCOCO~\cite{refcoco}{, } RefCOCO+~\cite{refcoco+}{, } RefCOCOg~\cite{refcoco+}{, } Ref-L4~\cite{ref-l4} 
                                , leaf, text width=40em
                        ]
                    ]
                    [
                       Fine-Grained \\ Perception  
                        [
                                FOCI~\cite{foci}{, } MMVP~\cite{mmvp}{, } LLVisionQA~\cite{q-bench}
                                , leaf, text width=40em
                        ]
                    ]
                    [
                       Video \\ Understanding  
                        [
                                Video-MME~\cite{video-mme}{, } MVBench~\cite{mvbench}{, } MLVU~\cite{mlvu}{, } LVBench~\cite{lvbench}{, }MMBench-Video~\cite{mmbench-video}{, } \\
                                Event-Bench~\cite{event-bench}{, }VN-Bench~\cite{vn-bench}{, }
                                EgoSchema~\cite{egoschema}{, }TempCompass~\cite{tempcompass}{, } \\
                                MSVD-QA~\cite{msvd-qa}{, }TGIF-QA~\cite{tgif-qa}{, }ActivityNet-QA~\cite{activity-net-qa}{, }MSRVTT-QA~\cite{msvd-qa}
                                , leaf, text width=40em
                        ]
                    ]
                ]
                [                
                    Model\\Self-Analysis\\(\S \ref{sec:benchmark_analysis})
                    [
                        Hallucination 
                        [
                                POPE~\cite{pope}{, } GAVIE~\cite{gavie}{, } M-HalDetect~\cite{m-haldetect}{, }
                                HaELM~\cite{haelm}{, }MMHal-Bench~\cite{mmhal-bench}{, }Bingo~\cite{bingo}{, }PhD~\cite{phd}{, }\\HallusionBench~\cite{hallusionbench}{, } 
                                AMBER~\cite{amber}
                                OpenChair~\cite{openchair}{, }MHaluBench~\cite{chen24unihd}{, }VHTest~\cite{vhtest}{, }\\
                                VALOR-Eval~\cite{valor-eval}{, }VideoHallucer~\cite{videohallucer}{, }
                                HQH~\cite{hqh}{, }R-Bench~\cite{r-bench}{, }VLind-Bench~\cite{vlind-bench}
                                , leaf, text width=40em
                        ]
                    ]
                    [
                        Bias
                        [
                                VLBiasBench~\cite{vlbiasbench}{, } Bingo~\cite{bingo}{, } MM-SpuBench~\cite{mm-spubench}
                                , leaf, text width=40em
                        ]
                    ]
                    [
                        Safety
                        [
                                VLLM-safety-bench~\cite{tu2023how}{, } MultiTrust~\cite{multitrust}{, } AttackVLM~\cite{attackvlm}{, }AdvDiffVLM~\cite{advdiffvlm}{, }MOSSBench~\cite{mossbench}
                                , leaf, text width=40em
                        ]
                    ]
                    [
                        Causation
                        [
                                CELLO~\cite{cello}
                                , leaf, text width=40em
                        ]
                    ]
                ]
                [                
                    Extended \\ Applications \\ (\S \ref{sec:benchmark_extended})
                    [
                        Medical Image 
                        [
                                VQA-RAD~\cite{vqa-rad}{, } PathVQA~\cite{pathvqa}{, } SLAKE~\cite{slake}{, }
                                PMC-VQA~\cite{pmc-vqa}{, } OmniMedVQA~\cite{omnimedvqa}{, } \\
                                RadBench~\cite{radbench}{, } GMAI-MMBench~\cite{chen2024gmai}
                                , leaf, text width=40em
                        ]
                    ]
                    [
                        Sentiment Analysis
                        [
                                EmoBench~\cite{emobench}{, } FABA-Bench~\cite{faba-bench}
                                , leaf, text width=40em
                        ]
                    ]
                    [
                        Remote Sensing
                        [
                                RSVQA~\cite{rsvqa}{, } RSIVQA~\cite{rsivqa}{, } VQA-TextRS~\cite{vqa-textrs}{, } RSVG~\cite{rsvg}{, } RSVGD~\cite{rsvgd}{, } RSIEval~\cite{rsieval}{, } \\
                                RRSIS-D~\cite{rrsis-d}{, } VRSBench~\cite{vrsbench}
                                , leaf, text width=40em
                        ]
                    ]
                    [
                        Agent
                        [
                                AppAgent~\cite{appagent}{, } Mobile-Eval~\cite{mobile-eval}{, } GPT4Tools~\cite{gpt4tools}
                                , leaf, text width=40em
                        ]
                    ]
                    [
                        Code Generation
                        [
                                CharMimic~\cite{chartmimic}{, } Web2Code~\cite{web2code}
                                , leaf, text width=40em
                        ]
                    ]
                    [
                        GUI
                        [
                                RefExp~\cite{wichers2018resolving}{, } Screen2Words~\cite{wang2021screen2words}{, } ScreenQA~\cite{hsiao2024screenqa} {, }  Rico-semantics~\cite{sunkara2022towards}{, } ScreenAI~\cite{baechler2024screenai}{, } 
                                Widget~\cite{li2020widget} 
                                , leaf, text width=40em
                        ]
                    ]
                    [
                        Transfer Capability
                        [
                                VLAA~\cite{tu2023how}{, } BenchLMM~\cite{cai2023benchlmm}{, } MMCBench~\cite{zhang2024benchmarking}{, }
                                , leaf, text width=40em
                        ]
                    ]
                    [
                        Knowledge Editing
                        [
                                MMEdit~\cite{cheng-etal-2023-edit}{, } VLKEB~\cite{huang2024vlkeb}{, } 
                                , leaf, text width=40em
                        ]
                    ]
                    [
                        Embodied AI
                        [
                                EQA~\cite{das2018embodied}{, } EPIC-KITCHENS~\cite{Damen2021PAMI}{, } Ego4D~\cite{grauman2022ego4d}{, } EMQA~\cite{datta2022episodic}{, } 
                                SQA3D~\cite{ma2022sqa3d}{, } 
                                MoTIF~\cite{burns2022dataset}{, }\\RH20T-P~\cite{chen2024rh20t}
                                EmbodiedScan~\cite{wang2024embodiedscan}
                                , leaf, text width=40em
                        ]
                    ]
                    [
                        Autonomous Driving
                        [
                                BDD-X~\cite{kim2018textual}{, } HAD~\cite{kim2019CVPR}{, } Talk2Car~\cite{deruyttere2019talk2car}{, } Rank2Tell~\cite{sachdeva2024rank2tell}{, } 
                                DRAMA~\cite{malla2023drama}{, } 
                                NuScenes-QA~\cite{qian2024nuscenes}{, }
                                \\
                                DriveLM~\cite{sima2023drivelm}{, }LingoQA~\cite{marcu2023lingoqa}{, }NuPrompt~\cite{wu2023language}{, }Reason2Drive~\cite{nie2023reason2drive}{, }MME-RealWorld~\cite{zhang2024mme}{, }\\ IDKB Dataset~\cite{lu2024lvlmsobtaindriverslicense}
                                , leaf, text width=40em
                        ]
                    ]
                ]
            ]
        \end{forest}
    }
    \caption{Categories of MLLM benchmarks. }
    \label{taxo_of_bench}
\end{figure*}

%% file: tables/tax_benchmarks.tex
\renewcommand{\arraystretch}{1.3} 
\begin{table*}
    \centering
    \caption{Statistics of representative MLLM benchmarks.}
    \label{tax-benchmarks}
    \resizebox{\textwidth}{!}{
    \begin{tabular}{l|c|c|c|c|c|c}
    \hline
    \textbf{Benchmarks} & \textbf{QA Pairs} & \textbf{Answer Type} & \textbf{Metrics} & \textbf{Categories}  & \textbf{Data Collection} & \textbf{Annotation}  \\
    \hline\hline
    VQA v2~\cite{goyal2017making} & 453.0k & Open-Ended & Deterministic & Comprehensive
     Evaluation  & Incorporating Samples From Existing Datasets  & Manually  \\ \hline 
    RefCOCO~\cite{refcoco}  & 10.0k & Open-Ended & Deterministic & Visual Grounding  & Incorporating Samples From Existing Datasets  & Manually  \\ \hline 
    RefCOCO+~\cite{refcoco}  & 10.0k & Open-Ended & Deterministic & Visual Grounding  & Incorporating Samples From Existing Datasets  & Manually  \\ \hline 
    RefCOCOg~\cite{refcoco}  & 14.0k & Open-Ended & Deterministic & Visual Grounding  & Incorporating Samples From Existing Datasets  & Manually  \\ \hline 
    EmbodiedQA~\cite{das2018embodied}  & 5.0k & Open-Ended & Deterministic & EmbodiedAI  & Incorporating Samples From Existing Datasets  &  Automatically Constructing  \\ \hline 
    TextVQA~\cite{singh2019towards}  & 5.7k & Open-Ended & Deterministic & OCR  & Incorporating Samples From Existing Datasets  & Manually  \\ \hline 
    Ego4D~\cite{grauman2022ego4d}  & 12.7k & Open-Ended & Deterministic & EmbodiedAI(Video)  & Gathering Data From the Internet  & Manually  \\ \hline 
    VizWiz~\cite{gurari2018vizwiz}  & 8.0k & Open-Ended & Deterministic & Comprehensive
     Evaluation  & Incorporating Samples From Existing Datasets  & Manually  \\ \hline 
    NLVR2~\cite{nlvr2}  & 2.0k & Yes-or-No & Deterministic & Multi-Image  & Gathering Data From the Internet  & Manually  \\ \hline 
    MME~\cite{fu2023mme}  & 2.3k & Yes-or-No & Deterministic & Comprehensive
     Evaluation  & Incorporating Samples From Existing Datasets  & Manually  \\ \hline 
    Video-MME~\cite{video-mme}  & 2.7k & Multi-Choice & Deterministic & Video Understanding  & Gathering Data From the Internet  & Manually  \\ \hline 
    MME-RealWorld~\cite{zhang2024mme}  & 29.4k & Multi-Choice & Deterministic & High Resolution  & Incorporating Samples From Existing Datasets  & Manually  \\ \hline 
    ScienceQA~\cite{lu2022learn}  & 21.0k & Multi-Choice & Deterministic &  Multidisciplinary  & Gathering Data From the Internet  & Manually  \\ \hline 
    DocVQA~\cite{docvqa}  & 50.0k & Open-Ended & Deterministic & Chart and Documentation  & Gathering Data From the Internet  & Manually  \\ \hline 
    BDD-X~\cite{kim2018textual}  & 6.9k & Open-Ended & Deterministic & Autonomous Driving(Video)  & Incorporating Samples From Existing Datasets  & Manually  \\ \hline 
    ActivityNet-QA~\cite{activity-net-qa}  &  58.0k & Open-Ended & Scoring & Video Understanding  & Gathering Data From the Internet  & Manually  \\ \hline 
    AI2D~\cite{ai2d}  & 4.5k & Multi-Choice & Deterministic & Chart and Documentation  & Gathering Data From the Internet  & Manually  \\ \hline 
    POPE~\cite{pope}  & 3.0k & Yes-or-No & Deterministic &  Hallucination  & Incorporating Samples From Existing Datasets  & Manually  \\ \hline 
    ChartQA~\cite{chartqa}  & 32.7k & Open-Ended & Deterministic & Chart and Documentation  & Gathering Data From the Internet  & Manually  \\ \hline 
    FigureQA~\cite{figureqa}  & 1.0M & Yes-or-No & Deterministic & Chart and Documentation  & Gathering Data From the Internet  &  Automatically Constructing  \\ \hline 
    OCR-VQA~\cite{mishra2019ocr}  & 100.0k & Open-Ended & Deterministic & OCR  & Incorporating Samples From Existing Datasets  &  Automatically Constructing  \\ \hline 
    MMBench~\cite{liu2023mmbench}  & 3.2k & Multi-Choice & Deterministic & Comprehensive Evaluation  & Gathering Data From the Internet  & Manually  \\ \hline 
    EPIC-KITCHENS~\cite{Damen2021PAMI}  & 10.9k & Open-Ended & Deterministic & Embodied AI(Video)  & Gathering Data From the Internet  & Manually  \\ \hline 
    MathVista~\cite{mathvista}  & 6.1k & Open-Ended/Multi-Choice & Deterministic & Mathematical  & Gathering Data From the Internet  & Manually  \\ \hline 
    MM-Vet~\cite{yu2024mm}  & 218 & Open-Ended & Scoring & Comprehensive Evaluation  & Gathering Data From the Internet  & Manually  \\ \hline 
    SEED-Bench~\cite{li2023seed}  & 19.2k & Multi-Choice & Deterministic & Comprehensive Evaluation  & Incorporating Samples From Existing Datasets  & Manually  \\ \hline 
    RSVQA~\cite{rsvqa}  & 299.6k & Open-Ended & Deterministic & Remote Sensing  & Gathering Data From the Internet  &  Automatically Constructing  \\ \hline 
    InfoVQA~\cite{mathew2022infographicvqa}  & 3.2k & Open-Ended & Deterministic & Chart and Documentation  & Gathering Data From the Internet  & Manually  \\ \hline 
    SLAKE~\cite{slake}  & 2.1k & Open-Ended/Multi-Choice & Deterministic & Medical Image  & Incorporating Samples From Existing Datasets  & Manually  \\ \hline 
    RSIVQA~\cite{rsivqa}  & 111.0k & Open-Ended & Deterministic & Remote Sensing  & Incorporating Samples From Existing Datasets  & Manually  \\ \hline 
    VisualMRC~\cite{visualmrc}  & 30.6k & Open-Ended & Deterministic & Chart and Documentation  & Gathering Data From the Internet  & Manually  \\ \hline 
    MMMU~\cite{mmmu}  & 11.5k & Open-Ended/Multi-Choice & Deterministic & Multidisciplinary  & Gathering Data From the Internet  & Manually  \\ \hline 
    PATHVQA~\cite{pathvqa}  & 6.0k & Open-Ended/Yes-or-No & Deterministic & Medical Image  & Gathering Data From the Internet  &  Automatically Constructing  \\ \hline 
    HAD~\cite{kim2019CVPR}  &  5.7k   & Open-Ended & Deterministic & Autonomous Driving(Video)  & Gathering Data From the Internet  & Manually  \\ \hline 
    Talk2Car~\cite{deruyttere2019talk2car}  & 850 & Open-Ended & Deterministic & Autonomous Driving(Video)  & Gathering Data From the Internet  & Manually  \\ \hline 
    LVLM-eHub~\cite{xu2023lvlm}  & 288.8k & Open-Ended/Multi-Choice & Deterministic & Comprehensive Evaluation  & Incorporating Samples From Existing Datasets  &  Automatically Constructing  \\ \hline 
    VQA-RAD~\cite{vqa-rad}  & 451 & Open-Ended/Yes-or-No & Deterministic & Medical Image  & Gathering Data From the Internet  &  Automatically Constructing  \\ \hline 
    MVBench~\cite{mvbench}  & 4.0k & Multi-Choice & Deterministic & Video Understanding  & Gathering Data From the Internet  &  Automatically Constructing  \\ \hline 
    LEAF-QA~\cite{leaf-qa}  & 1.6M & Open-Ended & Deterministic & Chart and Documentation  & Gathering Data From the Internet  &  Automatically Constructing  \\ \hline 
    GPT4tools~\cite{gpt4tools}  & 652 & Open-Ended & Deterministic & Agent  & Gathering Data From the Internet  & LLMs/MLLMs  \\ \hline 
    Screen2Words~\cite{wang2021screen2words}  & 21.5k & Open-Ended & Deterministic & GUI  & Gathering Data From the Internet  & Manually  \\ \hline 
    Widget Captioning~\cite{li2020widget}  & 12.3k & Open-Ended & Deterministic & GUI  & Gathering Data From the Internet  & Manually  \\ \hline 
    MMHAL-BENCH~\cite{mmhal-bench}  & 96 & Open-Ended & Scoring & Hallucination  & Incorporating Samples From Existing Datasets  & Manually  \\ \hline 
    AttackVLM~\cite{attackvlm}  & 50.0k & Open-Ended & Deterministic & Safety  & Incorporating Samples From Existing Datasets  & Manually  \\ \hline 
    PMC-VQA~\cite{pmc-vqa}  & 2.0k & Open-Ended/Multi-Choice & Deterministic & Medical Image  & Incorporating Samples From Existing Datasets  & LLMs/MLLMs  \\ \hline 
    EgoSchema~\cite{egoschema}  & 5.0k & Multi-Choice & Deterministic & Video Understanding  & Gathering Data From the Internet  & Manually  \\ \hline 
    M-HalDetect~\cite{m-haldetect}  & 3.2k & Open-Ended & Scoring & Hallucination  & Incorporating Samples From Existing Datasets  & Manually  \\ \hline 
    OCRBench~\cite{liu2024hidden}  & 1.0k & Open-Ended & Deterministic & OCR  & Incorporating Samples From Existing Datasets  & Manually  \\ \hline  
    GAVIE~\cite{gavie}  & 1.0k & Open-Ended & Scoring & Hallucination  & Incorporating Samples From Existing Datasets  & LLMs/MLLMs  \\ \hline 
    HallusionBench~\cite{hallusionbench}  & 1.1k & Yes-or-No & Scoring & Hallucination  & Modifying Existing Data  & Manually  \\ \hline 
    
    \end{tabular}
    }
\end{table*}

%% file: draft/031_bench_collect.tex
\section{Benchmark Construction}
\label{sec:bench_collect}
\input{tables/tax_how_to_eval}

In this section, we introduce the two main processes involved in constructing the benchmark: data collection and annotation as illustrated in~\cref{taxo_how_to_eval}. 
Our categorization is mainly based on whether human or model involvement is required during the collection or annotation process.

\subsection{Data Collection}

\textit{Incorporating samples from existing datasets} is a common and perhaps the most popular method of data collection. 
Compared to manually gathering data, directly using samples from public datasets is more cost-effective, but can also increase the risk of data leakage.
For instance, MMT-Bench~\cite{mmtbench} and SEED-Bench-2~\cite{li2023seed2} utilize data from public datasets, yet take annotation methods to reconstruct QA pairs to reduce the negative impact of data leakage.

\textit{Modifying existing data} is also an optional data collection method, particularly suitable for benchmarks that require specific data for evaluation. 
However, if manual modification is required, it can be a labor-intensive process. 
For example, VCR~\cite{zhang2024vcr} uses machine learning algorithms to occlude image captions to study the ability of restoring occluded text. 
MMCBench~\cite{zhang2024benchmarking} adds noise to data to study model robustness, while HallusionBench~\cite{hallusionbench} manually edites images to study model hallucination performance.

\textit{Gathering data from the Internet} is another commonly used data collection method. 
This approach effectively avoids overlapping with existing training data or benchmarks, but incurs higher labor costs and may lead to copyright issues, requiring careful filtering during the collection process. 
For example, ScienceQA~\cite{lu2022learn}, MMBench~\cite{liu2023mmbench}, and MME-RealWorld~\cite{zhang2024mme} gather massive Internet data for annotation.

\subsection{Annotation}
The annotation process can also be divided into three categories.
The most cost-effective method is to \textit{automatically construct QA pairs} by extracting relevant information from existing datasets using templates. 
This category primarily includes two approaches: the simplest one directly constructs from existing datasets using some selection criteria, and the other one rewrites existing annotations based on certain rules. 
For the former, MM-Vet~\cite{yu2024mm} includes some questions derived from its original annotations, and LVLM-eHub~\cite{xu2023lvlm} uses entirely annotations from existing datasets. 
MathVista~\cite{mathvista} selects samples that meet specific requirements from various benchmarks to create a new benchmark. 
The annotation quality in these cases largely depends on the quality of the existing benchmarks and may have overlap with other benchmarks. 
OCR-VQA~\cite{mishra2019ocr} designs fixed-genre questions to generate corresponding questions and answers for each image, while EQA~\cite{das2018embodied} uses programs for automatic annotation. This annotation method is feasible when the data format is relatively uniform.

\textit{Prompting LLMs or MLLMs to generate QA pairs} is currently a popular annotation method. 
As the performance of LLMs/MLLMs improves, using them for annotation combined with subsequent human review can yield a reasonably high-quality benchmark. 
Benchmarks like MMStar~\cite{chen2024we}, Seed-Bench~\cite{li2023seed}, MMT-Bench~\cite{mmtbench}, and SEED-Bench~\cite{li2023seed} have adopted this approach. 
This annotation process is inherently limited by the performance of the LLMs/MLLMs used. 
For instance, in MME-RealWorld~\cite{zhang2024mme}, the best-performing model, Qwen2-VL, achieved only 40\% accuracy in some tasks, indicating that relying on models inevitably introduces significant noise, which compromises the quality of the annotations.

\textit{Manual annotation} generally incurs the highest cost but ensures better quality. 
Benchmarks like VQA v2~\cite{goyal2017making}, VizWiz~\cite{gurari2018vizwiz}, and TextVQA~\cite{singh2019towards} are annotated by Amazon Mechanical Turk workers. 
MMBench~\cite{liu2023mmbench}, MME~\cite{fu2023mme}, VideoMME~\cite{fu2024video}, and MME-RealWorld~\cite{zhang2024mme} are also benchmarks that are purely manually annotated. 
However, due to the high labor costs, the data scale of these benchmarks is usually limited. 
To date, the largest purely manually annotated dataset, MME-RealWorld~\cite{zhang2024mme}, employed 32 annotators and contains 29K QA pairs.

\subsection{Common Challenges and Future Trends in Benchmark Construction}
In constructing benchmarks for MLLMs, several important considerations must be taken into account to ensure that the evaluation truly reflects the model’s capabilities. 
These issues, if overlooked, can lead to misleading conclusions about a model's performance.

\textbf{Multiple-Choice Question Leakage}.
Many benchmarks evaluate models using multiple-choice questions (MCQs)~\cite{fu2023mme,liu2023mmbench,zhang2024mme}, which are easier to automate in terms of evaluation and statistical analysis. 
However, this format introduces the possibility of models guessing the correct answer without truly understanding the question or image. 
The structure of multiple-choice questions, by nature, may inadvertently provide clues within the answer options, allowing models to ``game'' the system rather than demonstrating genuine reasoning capabilities. 

To address this, some benchmarks require models to provide reasoning steps along with their answers. This ensures that the model demonstrates a true understanding of the question and its components, as seen in benchmarks like ScienceQA~\cite{lu2022learn} and MMEvalPro~\cite{mmevalpro}. 
These benchmarks force the model to show how it arrived at the answer, rather than merely selecting an option. 
In addition, some benchmarks have designed more challenging or deceptive options to avoid trivial answers, making it harder for models to succeed without robust reasoning. 
Thus, the format of the questions should be carefully selected based on the evaluation scenario, ensuring that the model's ability to reason, not just guess, is properly evaluated.

\textbf{Data Leakage}.
Data leakage happens when models are evaluated using data they have already encountered during training.  
The issue becomes worse when benchmarks are built from existing academic datasets, where training and evaluation sets can overlap. 
To this end, it is vital to minimize overlap with public datasets during the benchmark construction, especially those used for training~\cite{fu2023mme}.
Using new or carefully selected data can lower the risk of leakage. 
Besides, techniques like deduplication and thorough data checks can help spot and remove cases where the same or very similar examples are used in both training and testing.

\textbf{Vision-Centric Evaluation}.
One significant issue with many existing benchmarks is that models can answer questions correctly without even processing the visual input, relying solely on the accompanying text. 
This undermines the purpose of evaluating MLLMs in multimodal tasks. 
In benchmarks like MMMU~\cite{mmmu}, it is found that many of the questions could be answered without understanding or even viewing the image, reducing the reliance on the vision modality. 
To counter this, new benchmarks such as Video-MME~\cite{fu2024video}, MMEvalPro~\cite{mmevalpro}, and CV-Bench~\cite{tong2024cambrian} have been designed with vision-centric tasks, where the visual content is essential for answering the questions. These benchmarks focus on ensuring that the model cannot answer the question solely based on text and must integrate visual information to provide accurate responses. For example, MMEvalPro~\cite{mmevalpro} uses triplet designs (a combination of an image, a question, and deliberately confusing options) to ensure that the MLLM has thoroughly processed the visual input. When constructing new benchmarks, it is essential to carefully curate questions where the image significantly impacts the answer, ensuring that the model genuinely understands and utilizes visual content.

\textbf{Benchmark Diversity and Sample Size}.
As MLLMs grow more capable, the complexity and variety of tasks they are evaluated on need evolve accordingly. 
Simple tasks are no longer sufficient to expose the limitations of these models. 
Additionally, benchmarks with small number run the risk of producing unrobust evaluation results. 
The high variance can lead to unreliable conclusions about model performance. 
As such, the representative MME-RealWorld~\cite{zhang2024mme} presents challenging real-world scenario oriented tasks, and increases the size of QA pairs to 29K, making it the largest manually annotated benchmarks to date.

%% file: tables/tax_how_to_eval.tex
\tikzstyle{my-box}=[
    rectangle,
    draw=hidden-draw,
    rounded corners,
    text opacity=1,
    minimum height=1.5em,
    minimum width=5em,
    inner sep=2pt,
    align=center,
    fill opacity=.5,
    line width=0.8pt,
]
\tikzstyle{leaf}=[my-box, minimum height=1.5em,
    fill=hidden-pink!80, text=black, align=left,font=\footnotesize,
    inner xsep=2pt,
    inner ysep=4pt,
    line width=0.8pt,
]
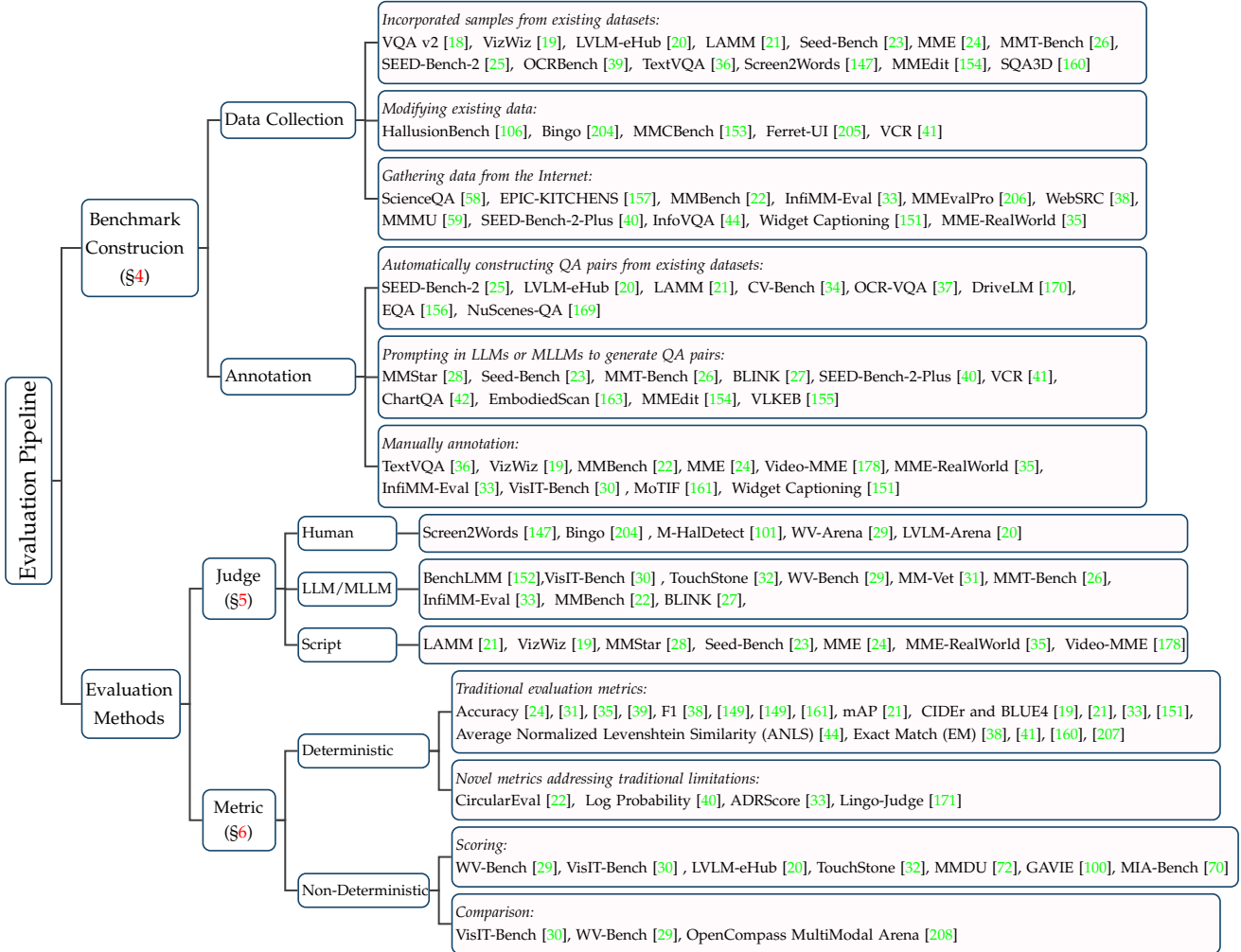
\begin{figure*}[!th]
    \centering
    \resizebox{0.96\textwidth}{!}{
        \begin{forest}
            forked edges,
            for tree={
                grow=east,
                reversed=true,
                anchor=base west,
                parent anchor=east,
                child anchor=west,
                base=left,
                font=\large,
                rectangle,
                draw=hidden-draw,
                rounded corners,
                align=left,
                minimum width=4em,
                edge+={darkgray, line width=1pt},
                s sep=3pt,
                inner xsep=2pt,
                inner ysep=3pt,
                line width=0.8pt,
                ver/.style={rotate=90, child anchor=north, parent anchor=south, anchor=center},
            },
            where level=1{text width=6em,align=center,font=\normalsize,}{},
            where level=2{text width=7em,align=center,font=\small,}{},
            where level=3{text width=6.8em,font=\footnotesize,}{},
            where level=4{text width=5em,font=\tiny,}{},
            [
                Evaluation Pipeline, ver
                [
                    Benchmark \\Construcion \\ (\S \ref{sec:bench_collect})
                    [
                        Data Collection 
                        [
                                \textit{Incorporated samples from existing datasets:}\\ VQA v2~\cite{goyal2017making}{, } VizWiz~\cite{gurari2018vizwiz}{, } LVLM-eHub~\cite{xu2023lvlm}{, } LAMM~\cite{yin2024lamm}{, } 
                                Seed-Bench~\cite{li2023seed}{, }MME~\cite{fu2023mme}{, } MMT-Bench~\cite{mmtbench}{, } \\
                                SEED-Bench-2~\cite{li2023seed2}{, } 
                                OCRBench~\cite{liu2024hidden}{, }
                                TextVQA~\cite{singh2019towards}{, }Screen2Words~\cite{wang2021screen2words}{, } MMEdit~\cite{cheng-etal-2023-edit}{, } 
                                SQA3D~\cite{ma2022sqa3d} 
                                , leaf, text width=42em
                        ]
                        [
                                \textit{Modifying existing data:} \\HallusionBench~\cite{hallusionbench}{, } Bingo~\cite{cui2023holistic}{, } MMCBench~\cite{zhang2024benchmarking}{, }
                                Ferret-UI~\cite{you2023ferret}{, } 
                                VCR~\cite{zhang2024vcr}
                                , leaf, text width=42em
                        ]
                        [
                                \textit{Gathering data from the Internet:}\\ScienceQA~\cite{lu2022learn}{, } EPIC-KITCHENS~\cite{Damen2021PAMI}{, } MMBench~\cite{liu2023mmbench}{, } InfiMM-Eval~\cite{han2023coremm}{, }MMEvalPro~\cite{mmevalpro}{, }
                                WebSRC~\cite{chen2021websrc}{, }\\
                                 MMMU~\cite{mmmu}{, }
                                SEED-Bench-2-Plus~\cite{li2024seed2plus}{, }InfoVQA~\cite{mathew2022infographicvqa}{, }
                                Widget Captioning~\cite{li2020widget}{, } MME-RealWorld~\cite{zhang2024mme}
                                , leaf, text width=42em
                        ]
                    ]
                    [
                       Annotation  
                       [
                            \textit{Automatically constructing QA pairs from existing datasets:}\\
                                SEED-Bench-2~\cite{li2023seed2}{, } LVLM-eHub~\cite{xu2023lvlm}{, } LAMM~\cite{yin2024lamm}{, } CV-Bench~\cite{tong2024cambrian}{, }OCR-VQA~\cite{mishra2019ocr}{, } 
                                DriveLM~\cite{sima2023drivelm}{, }\\
                                EQA~\cite{das2018embodied}{, } 
                                NuScenes-QA~\cite{qian2024nuscenes}
                                , leaf, text width=42em
                        ]             
                        [       
                            \textit{Prompting in LLMs or MLLMs to generate QA pairs:}\\
                                MMStar~\cite{chen2024we}{, } Seed-Bench~\cite{li2023seed}{, }  MMT-Bench~\cite{mmtbench}{, }  BLINK~\cite{fu2024blink}{, }SEED-Bench-2-Plus~\cite{li2024seed2plus}{, }VCR~\cite{zhang2024vcr}{, }
                                 \\
                                  ChartQA~\cite{chartqa}{, }  EmbodiedScan~\cite{wang2024embodiedscan}{, } MMEdit~\cite{cheng-etal-2023-edit}{, } VLKEB~\cite{huang2024vlkeb}
                                , leaf, text width=42em
                        ]       
                        [       
                            \textit{Manually annotation:}\\
                               TextVQA~\cite{singh2019towards}{, } VizWiz~\cite{gurari2018vizwiz}{, }MMBench~\cite{liu2023mmbench}{, }MME~\cite{fu2023mme}{, }Video-MME~\cite{fu2024video}{, }MME-RealWorld~\cite{zhang2024mme}{, }
                               \\
                               InfiMM-Eval~\cite{han2023coremm}{, }VisIT-Bench~\cite{bitton2023visit} {, }MoTIF~\cite{burns2022dataset}{, }
                                 Widget Captioning~\cite{li2020widget}  
                                , leaf, text width=42em
                        ]
                    ]
                ]
                [                
                    Evaluation \\ Methods, text width=5em
                    [
                        Judge\\ (\S \ref{sec:bench_eval_judge}), text width=2.5em
                        [
                            Human, text width=5em
                            [
                                   Screen2Words~\cite{wang2021screen2words}{, }Bingo~\cite{cui2023holistic} {, }M-HalDetect~\cite{m-haldetect}{, }WV-Arena~\cite{lu2024wildvision}{, }LVLM-Arena~\cite{xu2023lvlm}
                                    , leaf, text width=42em
                            ]
                        ]
                        [
                            LLM/MLLM, text width=5em
                            [
                                    BenchLMM~\cite{cai2023benchlmm}{,}VisIT-Bench~\cite{bitton2023visit} {, }TouchStone~\cite{bai2023touchstone}{, }WV-Bench~\cite{lu2024wildvision}{, }MM-Vet~\cite{yu2024mm}{, }MMT-Bench~\cite{mmtbench}{, } \\
                                    InfiMM-Eval~\cite{han2023coremm}{, } MMBench~\cite{liu2023mmbench}{, }BLINK~\cite{fu2024blink}{, }
                                    , leaf, text width=42em
                            ]
                        ]
                        [
                            Script, text width=5em
                            [
                                   LAMM~\cite{yin2024lamm}{, } VizWiz~\cite{gurari2018vizwiz}{, }MMStar~\cite{chen2024we}{, } Seed-Bench~\cite{li2023seed}{, }MME~\cite{fu2023mme}{, } MME-RealWorld~\cite{zhang2024mme}{, } Video-MME~\cite{fu2024video}
                                    , leaf, text width=42em
                            ]
                        ]
                    ]
                    [
                        Metric \\(\S \ref{sec:bench_eval_metric}),text width=2.8em
                        [
                            Deterministic
                            [\textit{Traditional evaluation metrics:}\\
                                    Accuracy~\cite{fu2023mme,zhang2024mme,yu2024mm,liu2024hidden}{, }F1~\cite{sunkara2022towards,chen2021websrc,sunkara2022towards,burns2022dataset}{, }mAP~\cite{yin2024lamm}{, } CIDEr and BLUE4~\cite{han2023coremm,gurari2018vizwiz,yin2024lamm,li2020widget}{, }\\
                                    Average Normalized Levenshtein Similarity (ANLS)~\cite{mathew2022infographicvqa}{,} Exact Match (EM)~\cite{chen2021websrc,li2024fakebench,zhang2024vcr,ma2022sqa3d}
                                    , leaf, text width=42em
                            ]
                            [\textit{Novel metrics addressing traditional limitations:}\\
                                     CircularEval~\cite{liu2023mmbench}{, } Log Probability~\cite{li2024seed2plus}{, }ADRScore~\cite{han2023coremm}{, }Lingo-Judge~\cite{marcu2023lingoqa}
                                    , leaf, text width=42em
                            ]
                        ]
                        [
                            Non-Deterministic
                            [\textit{Scoring:}\\
                                    WV-Bench~\cite{lu2024wildvision}{, }VisIT-Bench~\cite{bitton2023visit} {, }LVLM-eHub~\cite{xu2023lvlm}{, }TouchStone~\cite{bai2023touchstone}{, }MMDU~\cite{mmdu}{, }GAVIE~\cite{gavie}{, }MIA-Bench~\cite{MIA-Bench}
                                    , leaf, text width=43em
                            ]
                            [\textit{Comparison:}\\
                                    VisIT-Bench~\cite{bitton2023visit}{, }WV-Bench~\cite{lu2024wildvision}{, }OpenCompass MultiModal Arena~\cite{multimodal-arena}
                                    , leaf, text width=42em
                            ]
                        ]
                    ]
                ]
            ]
        \end{forest}
    }
    \caption{Evaluation pipeline of MLLMs. }
    \label{taxo_how_to_eval}
\end{figure*}

%% file: draft/032_bench_eval.tex
\section{Evaluation Judge} \label{sec:bench_eval_judge}
In this section, we introduce three evaluation strategies: human evaluation, LLM/MLLM-based evaluation, and script-based evaluation. 
The cost of these strategies decrease in the aforementioned order, though each comes with its own set of advantages and disadvantages.

\subsection{Human Evaluation}
Human evaluation of model response is considered the most effective method, as the ultimate goal of MLLMs is for human use. 
For instance, Screen2Words~\cite{wang2021screen2words} conducts a Mechanical Turk study to ask humans to assess the quality of generated screen summaries and validate how automatic metrics correlate with human judgment. 
Bingo~\cite{bingo} employs human annotators to evaluate the accuracy of GPT-4V's responses to analyze the model biases. 
M-HalDetect~\cite{m-haldetect} uses human evaluation to assess hallucination rates, demonstrating that human evaluation is more accurate than model-based assessments. 
WV-Arena~\cite{lu2024wildvision} adopts a human voting method to score models and uses Elo rating to compare multiple models. 
However, incorporating human evaluation undoubtedly increases time expenditure and labor costs. 
Besides, if the number of evaluators is small, there is a concern that individual preferences might influence the scores. 

\subsection{LLM/MLLM-based Evaluation}
LLM/MLLM-based evaluation methods can be categorized into two types based on the degree of model involvement:
1) \textit{Shallow Model Involvement}. In these scenarios, LLMs/MLLMs are only partially involved in the evaluation process, such as finding the most matching answer for a responsed string. For instance, MMbench~\cite{liu2023mmbench} and BLINK~\cite{fu2024blink} use GPT-4 and GPT-3.5-turbo respectively as choice extractors. This method is useful to MLLMs that are not strong in instruction following, providing a flexible alternative to match predictions and correct answers.
2) \textit{Full Model Responsibility}. This approach is primarily used for open-ended tasks where the format and content of correct answers are not fixed. In such cases, LLMs/MLLMs can compare reference answers with generated answers or score directly. For example, MM-Vet~\cite{yu2024mm} leverages GPT-4 to assist in evaluation, with GPT-4 automatically generating scores for each sample based on the input question, ground truth, and model output. Similarly, TouchStone~\cite{bai2023touchstone} and LLaVA-bench~\cite{liu2024visual} uses GPT-4 to directly compare generated answers with reference answers.

Incorporating other models in the evaluation process reduces human labor. However, this method is plagued with systematic biases~\cite{wang2023large}, such as sensitivity to the ordering of responses. Moreover, the evaluation results are significantly constrained by the capabilities of the LLMs/MLLMs themselves~\cite{zhang2024mme}. 
There are times when different LLMs result in completely different evaluation results.

\subsection{Script-based Evaluation}
Script-based evaluation methods are simpler and are commonly used in benchmarks based on multiple-choice or ``Yes-or-No'' questions. 
These evaluations compare results according to predefined rules. 
The MME series~\cite{fu2023mme,video-mme,zhang2024mme} adopt this approach by first performing regular expression matching on the output results to find the generated options, and then directly comparing the matched results with the ground truth.
For example, for a multiple-choice question, the output of MLLM is ``The answer is A'', and the ground true answer is ``B''. 
The role of the script is to extract ``A'' and compare it with ``B'' to determine whether it is right or wrong.

This method evaluates quickly but also has certain drawbacks. 
The final accuracy is heavily dependent on the effectiveness of the regular expression matching. If a model has poor instruction-following capabilities and the outputs are messy, this evaluation method may fail. 
Therefore, when using this evaluation method, it is crucial to construct appropriate prompts that make MLLMs output regular.

\section{Evaluation Metric} \label{sec:bench_eval_metric}
In this section, we introduce two primary categories of evaluation metrics: deterministic metrics and non-deterministic metrics. 
The distinction lies in whether the evaluation criteria are deterministic, that is, whether a definitive value can be reached by comparing the generated output with the ground truth.

\subsection{Deterministic Metrics}
 
These metrics typically rely on standardized evaluation tools, enabling objective assessment with minimal human intervention. 
Compared to subjective human evaluations, deterministic metrics offer the benefits of saving time, reducing bias, and ensuring consistency across different assessments. 
We have categorized the key deterministic metrics based on their prevalence in existing literature as follows:

\subsubsection{Traditional Evaluation Metrics}
\begin{itemize}
    \item \textit{Exact Match (EM)~\cite{chen2021websrc,li2024fakebench,zhang2024vcr,ma2022sqa3d}}: This metric determines whether the model's output is an exact replica of the ground truth, making it a straightforward measure in many tasks.
    \item \textit{Accuracy~\cite{fu2023mme,zhang2024mme,fu2024video,tang2024mtvqa,tong2024cambrian}}: It is widely used in tasks like multiple-choice QA, assessing how often the model's output option matches the correct answer.
    \item \textit{F1 Score~\cite{hsiao2024screenqa,baechler2024screenai,burns2022dataset,sachdeva2024rank2tell}}: It is particularly useful in binary classification tasks, which balances precision and recall to provide a comprehensive measure of the performance.
    \item \textit{mean Average Precision (mAP)~\cite{yin2024lamm,wichers2018resolving,sunkara2022towards,you2024ferret}}: This metric is crucial for evaluating the grounding abilities, such as accurately identifying and localizing elements within an image.
    \item  \textit{CIDEr, BLEU4~\cite{gurari2018vizwiz,li2020widget,wang2021screen2words,nie2023reason2drive}, and ANLS~\cite{baechler2024screenai}}: These metrics evaluate the similarity between generated text and reference text, with CIDEr and BLEU focusing on n-gram overlap, and ANLS (Average Normalized Levenshtein Similarity) assessing the minimum number of edits needed to convert one sequence into another, particularly for open-ended answers.
    \item \textit{Task-Specific Metrics}: Tailored metrics are employed in specialized fields. For instance, in autonomous driving, metrics like absolute error and correlation distance~\cite{kim2018textual,kim2019CVPR} are used to predict factors like acceleration and steering angles. For object trajectory prediction, Average Multiple Object Tracking Precision (AMOTA)~\cite{wu2023language} is commonly used, while in red-teaming scenarios, metrics such as Attack Success Rate (ASR)~\cite{tu2023how} are prevalent.
\end{itemize}

\subsubsection{Novel Metrics Addressing Traditional Limitations}
While traditional metrics are valued for their simplicity and clarity, they can fall short in capturing the nuances of complex tasks. 
To address these limitations, researchers have developed more sophisticated metrics, particularly for evaluating tasks that involve multi-step reasoning. 
Some of these innovative metrics include:
\begin{itemize}
    \item \textit{CircularEval~\cite{liu2023mmbench}}: This metric is designed to counteract the bias that some models exhibit towards specific options in multiple-choice questions. CircularEval requires the model to correctly answer a question after multiple random shufflings of the choices, ensuring a more reliable evaluation of its understanding.
    \item \textit{Log Probability~\cite{li2023seed2}}: In situations where a model's ability to follow instructions is lacking, it may generate irrelevant content rather than directly answering the question. This metric assesses the log probability of the first generated token, and the option that produces the highest probability is considered the model's choice.
    \item \textit{ADRScore~\cite{han2023coremm}}: Traditional text generation metrics like BLEU and CIDEr often evaluate outputs holistically, potentially overlooking the logical progression of reasoning steps. ADRScore is an aggregated metric specifically designed to assess the performance of models in tasks requiring chain-of-thought reasoning.
    \item \textit{Lingo-Judge~\cite{marcu2023lingoqa}}: This metric involves training a classifier that evaluates the question, human response, and model response to determine the correctness of the model's output. Studies have shown that Lingo-Judge correlates strongly with human ratings, making it a robust and reliable evaluation tool.
\end{itemize}

\subsection{Non-Deterministic Metrics} 
Non-deterministic metrics encompass the evaluation methods that the model outputs are assessed either by other models or by human experts. 
There are two primary evaluation approaches within this category, including scoring and direct comparison.

\subsubsection{Scoring}
Scoring is a commonly employed strategy for open-ended tasks. 
In this approach, the model or human judges assign scores based on provided requirements. 
For instance, in LLaVA-Bench~\cite{liu2024visual}, the authors utilize detailed prompts, instructing GPT-4 to ``Please rate the helpfulness, relevance, accuracy, and level of detail of the responses.'' 
Each assistant then receives an overall score on a scale from 1 to 10. 
Similarly, MIA-Bench~\cite{qian2024mia} adopts GPT-4o to score the responses of MLLMs for each instruction, returning a total score between 0 and 10. 
Additionally, Bingo~\cite{cui2023holistic} uses human annotators to evaluate the accuracy of GPT-4V’s responses, assigning a score of 1 for correct answers and 0 for incorrect ones. 
The reliability of this scoring strategy depends on the model or expert and whether the researchers provided detailed prompts or user requirements, and the evaluation results of different models and experts may vary greatly. 
Moreover, some methods also rely on a reference text, such as in TouchStone~\cite{bai2023touchstone}, where the model compares a human-written caption with the one generated by the model and predicts the degree of hallucination.

\subsubsection{Comparison}
As opposed to scoring, direct comparison involves comparing the evaluated model’s outputs with optimal outputs using advanced models or experts. 
This approach is often considered more straightforward and stable than scoring. 
Common metrics include \textit{Win-rate}, which represents the proportion of victories in comparison tests with other models, and \textit{Elo ranking}, a rating method originally developed for chess that not only considers the win rate but also the strength of the opponents in each victory or defeat. 
Elo rankings are dynamically updated, adjusting scores based on the relative strength of opponents, thereby creating a more balanced rating system. 
For example, in VisIT-Bench~\cite{bitton2023visit}, the authors collect human preference-based comparison results to compile human-guided Elo rankings and win rates for the evaluated models.

\section{Evaluation Toolkit} \label{sec:bench_eval_toolkit}
\begin{figure*}[!th]
    \centering
    \includegraphics[width=1\textwidth]{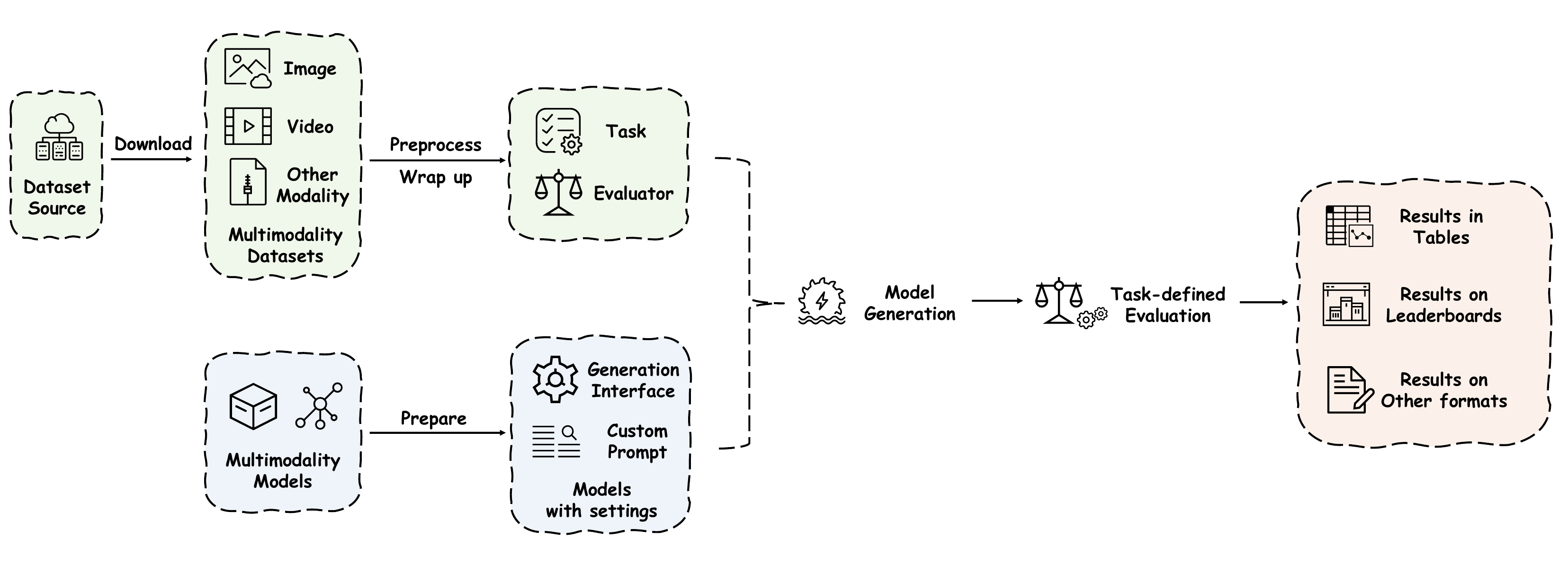}
    \caption{Major components and evaluation pipeline of the toolkit. By integrating various types of datasets and models, the evaluation toolkit facilitates the efficient acquisition and timely updating of assessment results, enabling comprehensive performance comparisons across models.}
    \label{fig:toolkit}
\end{figure*}
With the rapid iteration of MLLMs, it is essential for researchers to assess their capabilities across various aspects both during iterations and after new releases. 
However, the fragmented distribution of evaluation datasets, the tedious preparation work, and the potential for conflicts and mismatched results due to varying environmental requirements pose challenges to the rigorous evaluation of multimodal models on many benchmarks and the integration of results. 
The existence of toolkits facilitate the efficient integration of multiple datasets and models, enabling simultaneous evaluation across multiple datasets while simplifying environment configuration.

For the evaluation of LLMs, OpenAI introduces the Evals framework, which can test different dimensions of GPT models and allows for the creation of custom scripts for more comprehensive evaluations. 
OpenCompass~\cite{2023opencompass}, on the other hand, supports multiple evaluation datasets and models through a one-stop evaluation approach, offering versatility and high scalability. In the context of MLLMs, representative toolkits include VLMEvalKit~\cite{duan2024vlmevalkit} and LMMs-Eval~\cite{zhang2024lmms}. Additionally, the evaluation of agent capabilities and performance on medical tasks for MLLMs should not be overlooked. 
Toolkits such as agent-studio~\cite{zheng2024agentstudio} and MultiMedEval~\cite{royer2024multimedeval} have made it easier to assess these capabilities. 
\subsection{VLMEvalKit}
VLMEvalKit is a comprehensive, user-friendly, and easily extensible MLLM evaluation toolkit, which is designed to facilitate researchers to quickly evaluate the performance of existing MLLMs on multiple benchmarks. Currently, the codebase supports more than 70 different MLLMs, including proprietary APIs and open-source models, and more than 20 multimodal benchmarks covering a wide range of tasks and scenarios. 

For the vast array of datasets from different sources and formats, VLMEvalKit standardizes the preprocessing of data and stores multimodal content using paths or base64 encoding. 
Each evaluation benchmark is associated with a corresponding TSV file, which can be downloaded with a single click and verified for integrity using MD5 checksums. Each row in the TSV file represents a single evaluation sample, including \texttt{index}, \texttt{question}, \texttt{answer}, \texttt{image} or \texttt{image\_path} and \texttt{choices} for multi-choice questions. 
Each dataset class inherits from a corresponding base class, such as VQA, MCQ, Y/N, based on its question type and supports the \texttt{.build\_prompt()} interface. The evaluation scripts use this function to construct multimodal messages from evaluation samples, creating interleaved sequences of various modal contents.

VLMEvalKit implements a unified \texttt{.generate()} interface for different MLLMs. This interface accepts multimodal information, including multi-turn conversations with images, as input and returns a response string. 
The custom processing of multimodal information by the model is determined and constructed using the additional parameter \texttt{dataset\_name} of the \texttt{.generate()} interface and the model's own \texttt{.build\_prompt()} interface. 
It also supports differentiated adjustments to inference hyperparameters according to \texttt{dataset\_name}.

To accelerate multimodal inference, VLMEvalKit leverages Python's multiprocessing capabilities to support parallel inference with commercial APIs. 
It also fully utilizes computational resources to achieve multi-GPU distributed inference for open-source models. By temporarily storing results in \texttt{.pkl} files, the final inference results are complete and can be restored with minimal cost. For the final evaluation, VLMEvalKit additionally designs an LLM-based judge extractor to assist in answer matching for MCQ and Y/N question types. 
This extractor can be invoked through commercial APIs or deployed using LMDeploy.
For ongoing support of evaluation results, the VLMEvalKit team has established a public leaderboard\footnote{\url{https://rank.opencompass.org.cn/leaderboard-multimodal}}. This ensures that all evaluation results are openly accessible and reproducible, providing the community with an efficient means to assess and compare the performance of different models.

\subsection{LMMs-Eval}
LMMs-Eval is a unified and standardized benchmark for evaluating MLLMs, supporting over 50 tasks and more than a dozen models. By preprocessing datasets and recording model outputs, LMMs-Eval enables one-click evaluation across multiple tasks, thereby reducing the overhead associated with data collection and fragmented evaluation results. 
Additionally, LMMs-Eval establishes a unified framework encompassing various evaluation settings to achieve standardized and fair assessments. 
However, challenges remain, such as the high cost of comprehensive task evaluation and potential issues with evaluation dataset contamination.

To address this issue, the pruned evaluation toolkit, LMMs-Eval lite, has been proposed. By utilizing a greedy algorithm to solve the k-center problem, LMMs-Eval lite identifies a subset of benchmarks where the absolute scores and relative rankings of models are similar to those of the full set. 
Correlation calculations ensure that the selected subset maintains adequate testing capability. 
LMMs-Eval lite encompasses 6 task categories, including 15 datasets, and employs specific strategies to aggregate the scores of each subset. This allows researchers to efficiently evaluate model performance during the training phase.

Traditional evaluation benchmarks use fixed questions and answers for static assessment, which do not align with real-world usage scenarios. 
While benchmarks like Vibe-Eval~\cite{padlewski2024vibe} and LLaVA-Bench (Wilder)~\cite{llava-next} utilize real-world data to test model capabilities, the continuous updating of training data makes data contamination inevitable. 
LiveBench addresses this by periodically collecting real news from the internet and using powerful commercial multimodal models to construct QA pairs, forming a monthly evaluation dataset. 
This approach ensures the data remains authentic and minimizes contamination, while controlling the number of questions helps manage the cost of dataset construction and updates.

\subsection{MultiMedEval}
In the medical domain, LLaVA-Med~\cite{li2024llavamed} has been fine-tuned using data from PMC, effectively creating an assistant capable of multimodal medical question answering. 
Similarly, BioMedGPT~\cite{zhang2023biomedgpt} integrates multi-scale cross-modal biomedical data to establish a general-purpose MLLM capable of handling various tasks. However, existing medical evaluation benchmarks span multiple modalities, including X-ray, CT, general medical knowledge, and radiology, and encompass various tasks such as QA and summarization. 
This necessitates a toolkit to unify and simplify the evaluation methods across these benchmarks. 
Consequently, MultiMedEval~\cite{royer2024multimedeval} is developed to meet this need.

MultiMedEval encompasses 6 medical tasks, including natural language inference, report summarization, visual QA, and medical image classification, and covers 23 datasets from 11 different medical modalities. 
Researchers can easily install and use MultiMedEval via PyPi. After setting up the datasets to be evaluated, they only need to implement a callable \texttt{batcher} function for their models to initiate the evaluation and obtain the results in JSON format. Each call to this function will pass a list of prompts and return a corresponding list of decoded model responses for subsequent evaluation. Custom parameters such as data storage directories and batch size can be configured using the \texttt{SetupParams} and \texttt{EvalParams} classes.

\subsection{AgentStudio}
Enabling MLLMs to interact with the environment through external tools offers a practical and realistic assessment of their performance. 
Notably, virtual agents, which receive computer states and respond to instructions by invoking functions or manipulating software, have shown significant progress. 
However, these agents are developed for various domains, including gaming, online shopping, and web navigation, and there is a lack of a unified real-world setting and accompanying infrastructure to comprehensively evaluate their fundamental capabilities.

AgentStudio~\cite{zheng2024agentstudio} is a toolkit designed based on real-world environments, compatible with multiple operating systems and devices, encompassing the entire lifecycle of testing virtual agents. 
To address the previous inconsistencies in action spaces and observation spaces, AgentStudio provides a unified framework. It supports function calls and mouse/keyboard controls to manipulate any software and allows agents to use tools to access the internal structured state of programs like HTML. 
This avoids the limitations of single observation methods and promotes research on multimodal agent observation. 
AgentStudio also features a visual interface to monitor agent behavior and supports fully online, interactive environments, reflecting the complexity of real-world scenarios more accurately.

Additionally, AgentStudio offers an interactive annotation pipeline for labeling GUI-based data to evaluate the fundamental capabilities of models. 
It has been found that current advanced proprietary models, such as GPT-4V, still fall short in achieving robust and precise mouse action localization when acting as virtual agents and exhibit varied performance across different operating systems. AgentStudio also introduces a benchmark suite consisting of 77 real-world tasks, providing a comprehensive evaluation of existing agents' tool usage, compositional generalization, and other capabilities with three levels of increasing difficulty.

\subsection{Further Development}
Beyond the image modality, the development of MLLMs in video, audio, and other modalities has also heightened the need for comprehensive model evaluation. 
VLMEvalKit has taken the lead by incorporating several mainstream video understanding benchmarks and video MLLMs. 
However, issues remain, such as the lack of standardized video formats, differing frame selection methods across models, and the inability to perform integrated evaluations of video and audio. 
These aspects require further standardization and unification. Besides, there is ongoing demand within the community for evaluations of models in speech and 3D understanding, which could be a future direction for the development of comprehensive MLLM evaluation toolkits.

In more specific scenarios, MLLMs can currently interact as virtual agents. 
However, existing virtual agent evaluation toolkits still heavily rely on manual assessment and lack sufficient scenario diversity, necessitating further development. 
Furthermore, in fields such as medicine and embodied intelligence, there are issues with insufficient integration or a lack of evaluation toolkits, preventing unified performance assessments of various models. 
Therefore, it is essential to develop specialized evaluations alongside general ones to bring models closer to practical applications.

%% file: draft/040_future.tex
\section{Challenges and Future Directions}
\label{sec:future}

With the development of MLLMs, the need for comprehensive evaluation has gradually received increasing attention. 
While the academic and industrial communities have introduced over a hundred benchmarks for assessing these models, several challenges persist in the current evaluation landscape. 
Firstly, there is a notable lack of a universally accepted, standardized capability taxonomy, with existing evaluation sets often defining their own disparate ability dimensions. 
Secondly, current evaluation benchmarks exhibit gaps in their coverage of critical capabilities, particularly in areas such as instruction following, complex multi-modal reasoning, multi-turn dialogue experience, and creativity assessment. 
Thirdly, there is a dearth of task-specific evaluations for MLLMs, especially in commercially relevant domains such as invoice recognition, multi-modal knowledge base comprehension, and UI understanding and manipulation. 
Lastly, while existing multi-modal evaluation sets predominantly focus on image and video modalities, there remains a significant deficit in the assessment of capabilities related to audio and 3D representations. Addressing these challenges will be crucial for developing more robust and comprehensive evaluation methodologies for large multi-modal models in the future.

\subsection{Well-Defined Capability Taxonomy}
The rapid proliferation of multimodal benchmarks has rapidly expanded the breadth and depth of multimodal evaluation. 
These benchmarks typically define anywhere from a few to dozens of capability dimensions. 
However, there is a significant degree of overlap among these dimensions between different benchmarks. 
For example, capabilities such as OCR, celebrity recognition, and scene understanding are commonly found in benchmarks like MME~\cite{fu2023mme} and MMBench~\cite{liu2023mmbench}. 
The redundancy underscores the urgent need for an intricately developed, widely accepted capability taxonmomy within the multimodal domain to facilitate coherent progress. 
The design of such a taxonomy presents multiple avenues of exploration. 
Given that alignment with human cognition is a crucial feature of MLLMs, the taxonomy should be grounded in strong theoretical foundations, aligning with established research in psychology and cognitive science.

\subsection{Capability-Oriented Evaluation}
Despite rapid development, current MLLM evaluations remain not comprehensive enough, focusing primarily on assessing perception and reasoning abilities through objective questions~\cite{fu2023mme,mmmu,liu2023mmbench,mmtbench}. 
This creates a discernible gap between evaluation methodologies and real-world application scenarios. 
Moreover, optimizing models based on objective assessment results often leads developers to incorporate an abundance of objective question corpus during instruction tuning, 
potentially compromising the quality of actual dialogue experiences. 
Although subjective multimodal evaluation platforms such as WildVision~\cite{lu2024wildvision} and OpenCompass MultiModal Arena\footnote{\url{https://opencompass.org.cn/arena?type=multimodal}} have emerged, 
more research is needed to develop assessment methods that align more closely with practical usage scenarios. 
Current evaluation strategies largely rely on collecting or crafting specific questions to assess particular capabilities. 
However, complex multi-modal problems typically require the integration of multiple skills. 
For example, a chart-related question may involve OCR, spatial relationship recognition, reasoning, and calculations. 
The lack of a decoupled assessment~\cite{qiao2024prismframeworkdecouplingassessing} for these distinct capabilities represents a significant limitation in current evaluation frameworks. 
Furthermore, existing multimodal evaluations do not adequately cover crucial abilities, such as instruction following, with only a few benchmarks such as VisIT-Bench~\cite{bitton2023visit} and MIA-Bench~\cite{MIA-Bench} addressing this aspect. 
Multiturn dialogue, the primary mode of human interaction with multimodal models, remains a weakness for most current models, and the corresponding evaluations are still in their infancy, such as ConvBench~\cite{convbench} and MMDU~\cite{mmdu}. 
In the realm of complex multimodal reasoning, current evaluations predominantly focus on solving mathematical~\cite{mathvista,math-vision,mathverse} and examination problems~\cite{mmmu,cmmmu}, necessitating improvements in both difficulty and alignment with everyday use cases. 
Notably, the evaluation of multimodal creative tasks, a key application area for these models, such as text generation based on image and textual prompts, 
remains largely unexplored, highlighting a critical gap in the current evaluation landscape.

\subsection{Task-Oriented Evaluation}
As MLLM are still in the early stages of development, their business applications remain limited. 
Consequently, current evaluations predominantly focus on assessing fundamental capabilities rather than performance in real-world applications. 
Moving forward, it is crucial to develop evaluation frameworks that assess MLLM performance on specific tasks, particularly those with commercial value. 
Such tasks may include large-scale document processing, multimodal knowledge base comprehension, anomaly detection, and industrial visual inspection. 
When constructing task-specific evaluations for MLLMs, it is essential to consider not only the performance metrics, but also computational costs and inference speeds, comparing them against traditional computer vision-based methods such as OCR~\cite{du2020pp,kuang2021mmocr}, object detection~\cite{redmon2016you,he2017mask,ren2016faster,carion2020end}, action recognizers~\cite{wang2016temporal,feichtenhofer2019slowfast,duan2022revisiting}, to assess their practical applicability. 
Furthermore, a significant function of MLLMs lies in their potential in planning, interacting with environments as agents to solve complex problems~\cite{yang2023appagent,wang2024mobile}. 
Developing diverse virtual environments that allow MLLMs to interact and demonstrate their problem-solving capabilities as agents will likely become a critical component of future evaluations. 
Currently, evaluation efforts in this domain are in their early stages~\cite{xie2024osworld,rawles2024androidworld,koh2024visualwebarena}, indicating a promising area for future research and development in the field of multimodal AI assessment.

\subsection{Incorporating More Modalities}
Current evaluations of MLLMs primarily focus on image and video modalities, with limited attention given to other modalities. 
In the audio domain, models like Qwen-Audio~\cite{chu2023qwen} evaluate the audio-related capability on traditional audio tasks such as Automatic Speech Recongition~\cite{panayotov2015librispeech}, Speech-to-Text Translation~\cite{wang2021covost}, Acoustic Scene Classification~\cite{drossos2020clotho}, and Vocal Sound Classification~\cite{gong2022vocalsound}. 
However, there remains a notable gap in assessing capabilities related to speech meta-information recognition, such as accent recognition, emotion detection, and beyond.
The field of 3D modality evaluation is also in its infancy, with works like ScanRefer~\cite{chen2020scanrefer} and ScanReason~\cite{zhu2024empowering} representing early efforts. 
Furthermore, with the emergence of omni MLLMs like GPT-4o, Gemini, and VITA~\cite{fu2024vita}, there is an urgent need to develop evaluation frameworks that can assess a model's simultaneous perception and cross-modal reasoning capabilities across multiple modalities. 
This evolving landscape underscores the importance of expanding and diversifying evaluation methodologies to keep pace with the rapidly advancing capabilities of MLLMs, ensuring comprehensive assessment across a wider range of modalities and their interactions.

%% file: draft/050_conclusion.tex
\section{Conclusion}
\label{sec:conclusion}
MLLMs are developing rapidly, and the evaluation benchmarks are escorting them.
This paper has presented a thorough survey of MLLM evaluation, focusing on four foundational dimensions: what capabilities are evaluated, how to construct benchmarks, how to assess performance, and where is the direction of the next benchmark. 
We hope that this survey will answer questions from researchers about MLLM evaluation, and help with the development of the evaluation benchmarks and the model itself.